\documentclass[5p,twocolumn,authoryear,nopreprintline]{elsarticle}
\usepackage[T1]{fontenc}
\usepackage[utf8]{inputenc}
\usepackage{amsmath,amssymb}
\usepackage{graphicx,booktabs,multirow,threeparttable,tabularx}
\usepackage{xcolor}
\DeclareUnicodeCharacter{2011}{\nobreakdash-}
\usepackage{bibunits}
\usepackage{hyperref}
\definecolor{manuscriptblue}{RGB}{0,80,158}
\hypersetup{colorlinks=true,citecolor=manuscriptblue,linkcolor=manuscriptblue,
  urlcolor=manuscriptblue,bookmarksdepth=2,
  pdfauthor={Jicheng Ma, Yunyan Yang, Juan Zhao, and Liang Zhao}}
\providecommand{\doi}[1]{doi:\ \href{https://doi.org/#1}{\nolinkurl{#1}}}
\journal{Journal of Molecular Graphics and Modelling}
\newcommand{\runinhead}[1]{\par\noindent\textbf{#1}\enspace\ignorespaces}
\hypersetup{pdftitle={CurvFlow-DTA: dual-graph discrete Ricci curvature flow for drug--target affinity prediction}}

\newcommand{\mypar}[1]{\noindent\textbf{#1}}
\newcounter{algorithm}
\makeatletter
\newcommand{\arxivbibnamespace}[1]{%
  \def\@extra@b@citeb{.#1}%
  \def\@extra@binfo{.#1}%
}
\newcommand{\arxivsupplementsetup}{%
  \global\let\@title\@empty
  \global\let\useelstitle\@empty
  \global\let\elsauthors\@empty
  \global\let\useauthors\@empty
  \global\let\elsprelimauthors\@empty
  \global\let\elsaddress\@empty
  \global\let\authorsep\@empty
  \global\let\prelimauthorsep\@empty
  \global\let\sep\@empty
  \global\let\eadsep\@empty
  \global\let\@cornotes\@empty
  \global\let\@corref\@empty
  \global\let\@fnmark\@empty
  \setcounter{cnote}{0}%
  \setcounter{affn}{0}%
  \providecommand{\theHaffn}{}%
  \renewcommand{\theHaffn}{SI.\arabic{affn}}%
  \providecommand{\theHcnote}{}%
  \renewcommand{\theHcnote}{SI.\arabic{cnote}}%
  \setcounter{section}{0}%
  \setcounter{subsection}{0}%
  \setcounter{subsubsection}{0}%
  \setcounter{table}{0}%
  \setcounter{figure}{0}%
  \setcounter{equation}{0}%
  \setcounter{algorithm}{0}%
  \renewcommand{\thetable}{S\arabic{table}}%
  \renewcommand{\thefigure}{S\arabic{figure}}%
  \renewcommand{\thesection}{S\arabic{section}}%
  \renewcommand{\thealgorithm}{S\arabic{algorithm}}%
  \renewcommand{\theequation}{S\arabic{equation}}%
  \renewcommand{\theHsection}{SI.\arabic{section}}%
  \renewcommand{\theHsubsection}{SI.\arabic{section}.\arabic{subsection}}%
  \renewcommand{\theHsubsubsection}{SI.\arabic{section}.\arabic{subsection}.\arabic{subsubsection}}%
  \renewcommand{\theHtable}{SI.\arabic{table}}%
  \renewcommand{\theHfigure}{SI.\arabic{figure}}%
  \renewcommand{\theHequation}{SI.\arabic{equation}}%
  \renewcommand{\theHfootnote}{SI.\arabic{Hfootnote}}%
  \providecommand{\theHalgorithm}{}%
  \renewcommand{\theHalgorithm}{SI.\arabic{algorithm}}%
}
\makeatother

\begin{document}
\begin{bibunit}[abbrvnat-jmgm]
\arxivbibnamespace{main}

\begin{frontmatter}
\title{CurvFlow-DTA: dual-graph discrete Ricci curvature flow for drug--target affinity prediction}
\author[a]{Jicheng Ma}
\author[a]{Yunyan Yang}
\author[a]{Juan Zhao}
\author[b]{Liang Zhao\corref{cor1}}
\address[a]{School of Mathematics, Renmin University of China, Beijing, 100872, China}
\address[b]{School of Mathematical Sciences, Key Laboratory of Mathematics and Complex Systems of MOE, Beijing Normal University, Beijing, 100875, China}
\cortext[cor1]{Corresponding author.}
\ead{liangzhao@bnu.edu.cn}
\begin{abstract}

Graph neural networks are widely used for drug--target affinity (DTA) prediction, and discrete Ricci curvature has recently been used to characterize molecular graph geometry. Existing curvature-aware DTA approaches mainly use static curvature on the drug graph while representing proteins primarily with sequence-derived features. This leaves pair-adaptive use of graph geometry underexplored, which may limit adaptation to unseen entities in cold-start settings relevant to practical screening.\\
We present CurvFlow-DTA, which replaces a single static curvature representation with weighted Forman curvature flow on both molecular and protein residue--residue contact graphs. A label-independent flow trajectory is precomputed for each entity, and a pair-conditioned selector determines the horizons read by a dual-branch Flow-GINE. A frozen ESM-2 supplies residue-level representations and contact scores used to construct the protein graph. Inference requires only SMILES strings and protein sequences, without a bound complex structure.\\
On Davis and KIBA, CurvFlow-DTA improves on the protocol-matched Ricci-GraphDTA baseline in every warm and cold-start setting. Warm-split mean squared error (MSE) decreases by $19.9\%$ on Davis and $18.9\%$ on KIBA. Across the six cold-start comparisons, MSE decreases by $14.3$--$27.4\%$, with higher concordance index (CI) throughout. Within our compiled set of literature baselines, CurvFlow-DTA achieves the lowest MSE on both warm benchmarks and across four out of six cold-start evaluation settings.
\end{abstract}
\end{frontmatter}

\section{Introduction}

Quantifying the binding affinity between small molecules and protein targets is a central task in drug discovery. Drug--target affinity (DTA) prediction supports virtual screening, lead optimization, and drug repurposing by providing a continuous measure of interaction strength that enables more fine-grained ranking than a binary interaction label. Experimental affinity measurements remain resource-intensive relative to the large chemical--target search space. Accurate computational affinity prediction narrows this gap by focusing experimental effort on the most promising candidates. The Davis kinase panel and the KIBA integrated bioactivity dataset have become widely used benchmarks for DTA prediction~\citep{davis2011comprehensive,tang2014making,ozturk2018deepdta}.

Early DTA approaches used drug and target similarities or engineered pair features, as in kernel-based regularized least squares~\citep{pahikkala2015toward} and the gradient-boosting model SimBoost~\citep{he2017simboost}. DeepDTA subsequently learned representations directly from SMILES strings and protein sequences with convolutional neural networks~\citep{ozturk2018deepdta}, while TransformerCPI introduced a Transformer-based multi-head attention mechanism to model compound--protein interaction features more explicitly~\citep{chen2020transformercpi}. GraphDTA replaced the string-based drug encoder with a molecular graph neural network while retaining a sequence-based protein encoder~\citep{nguyen2021graphdta}, whereas DGraphDTA further represented the protein using a predicted residue contact graph~\citep{jiang2020dgraphdta}. More recently, protein language models such as ESM-2 have provided residue-level representations containing rich sequence and structural information~\citep{rives2021biological,lin2023esm2}, and PGraphDTA explored protein language models together with predicted contact information for affinity prediction~\citep{bal2023pgraphdta}. These developments make it possible to construct informative representations on both sides of a drug--target pair without requiring an experimentally determined bound complex. Evaluation on entity-disjoint splits is correspondingly important for assessing prediction when drugs or targets have not been observed during training~\citep{pahikkala2015toward}.

Message passing in graph neural networks (GNNs) is influenced by the topology and geometry of the underlying graph. When information from many distant nodes must propagate through a limited set of connections, it can be compressed during message passing, a phenomenon commonly referred to as over-squashing~\citep{alon2021bottleneck}. Curvature-based analyses have linked negatively curved graph regions to such structural bottlenecks and motivated geometry-aware graph rewiring~\citep{topping2022oversquashing}. Among discrete notions of graph curvature, Forman--Ricci curvature provides an edge-based combinatorial measure that can be computed efficiently on weighted graphs~\citep{forman2003bochner,sreejith2016forman}. Forman curvature flows have also been used to evolve edge weights in complex networks~\citep{weber2017formanflow}, while other discrete Ricci-flow formulations have been studied for network analysis~\citep{ni2019community}. More recently, Graph Neural Ricci Flow introduced time-varying curvature into feature evolution on general attributed graphs~\citep{chen2025graphneuralricciflow}. Within DTA, Ricci-GraphDTA incorporates static Forman curvature into molecular message passing while representing the target from its amino-acid sequence~\citep{zheng2026riccigraphdta}. These studies motivate the use of graph curvature in representation learning, but the use of an evolving curvature-guided edge metric on both drug and protein graphs, with its effective scale selected for a particular drug--target pair, remains underexplored.

We therefore introduce CurvFlow-DTA, a dual-graph DTA framework that applies weighted Forman curvature flow to both the molecular graph and the protein contact graph. For each entity, a label-independent flow trajectory is precomputed once and reused across all pairs containing that entity. A pair-conditioned selector predicts separate flow horizons for the drug and protein branches, so that the geometric scale read from each trajectory can vary with the drug--target pair. The selected edge metrics are incorporated through parallel Flow-GINE encoders, whose pooled representations are subsequently fused for affinity prediction. Protein node representations and contact scores are obtained from a frozen ESM-2 model. Consequently, inference requires only a SMILES string and a protein sequence and does not require a bound drug--protein complex structure. The main contributions are:
\begin{itemize}
    \item \textbf{Dual-graph curvature flow.}
    We extend curvature-aware DTA from static curvature representations to weighted Forman curvature flow trajectories on both molecular and protein graphs. The flow evolves edge metrics while preserving graph topology, providing an ordered set of curvature-guided graph states for downstream message passing.
    \item \textbf{Pair-conditioned flow scale selection.}
    A lightweight pair-conditioned selector uses information from both entities to determine separate flow horizons for the drug and protein branches. Flow-GINE then reads trajectory states at pair-dependent flow scales from trajectories that are precomputed independently of affinity labels.
    \item \textbf{Controlled evaluation under warm and cold-start settings.}
    We evaluate CurvFlow-DTA under warm, cold-drug, cold-target, and dual-cold settings on Davis and KIBA, together with matched controls that separately test static curvature, flow evolution, pair conditioning, edge--trajectory correspondence, and drug- versus protein-side flow. The resulting ablations consistently support the contributions of curvature evolution, pair-conditioned horizon selection, and dual-graph flow across the evaluated settings.
\end{itemize}

On Davis and KIBA, CurvFlow-DTA improves on the protocol-matched curvature baseline under both warm and cold-start evaluation. Comparisons against a range of published DTA baselines further show that CurvFlow-DTA is a competitive DTA predictor, with low regression error across both warm and cold evaluation settings. We conduct matched ablation experiments across these settings. We also examine protein-branch attributions on two held-out kinase complexes to check their agreement with crystallographic binding pockets.

\section{Related work}

\runinhead{Protein-aware DTA without bound complexes.}
DTA models increasingly enrich sequence-based protein encoders with predicted structural information and pretrained protein representations. DGraphDTA represents both entities as graphs, constructing the protein graph from a predicted residue contact map~\citep{jiang2020dgraphdta}. Interaction-aware architectures allow information from one entity to influence the representation of the other. GEFA uses an early-fusion graph-in-graph architecture~\citep{nguyen2022gefa}, whereas FusionDTA combines pretrained protein representations with attention-based feature aggregation~\citep{yuan2022fusiondta}. NHGNN-DTA integrates drug and protein graphs into a hybrid graph~\citep{he2023nhgnndta}. CSCo-DTA connects molecular representations with drug--target network representations through cross-scale graph contrastive learning~\citep{wang2024cscodta}. PGraphDTA combines protein language model representations with predicted contact maps~\citep{bal2023pgraphdta}, while KANPM-DTA constructs an ESM-guided protein graph using residue representations and contact information~\citep{rakib2026kanpmdta}. LLMDTA uses pretrained molecular and protein features for cold-start affinity prediction~\citep{tang2025llmdta}. PMMR combines pretrained protein and molecular representations with drug-graph features~\citep{ouyang2025pmmr}. PCIM-DTA derives a pair-specific condition vector to modulate interaction representations and affinity prediction~\citep{zhou2026pcimdta}. These studies demonstrate the value of predicted protein structure, pretrained representations, and cross-entity interaction modeling in DTA prediction. CurvFlow-DTA addresses the same prediction task, but focuses on a different aspect. It evolves edge metrics on both entity graphs, caches trajectories independently of their interaction partners, and selects the flow horizon read by each branch for a particular pair.

\runinhead{Graph curvature and curvature flows.}
Discrete Ricci curvature provides a practical tool to characterize connectivity and motivates geometry-aware graph learning~\citep{topping2022oversquashing}. Forman--Ricci curvature offers an efficiently computable edge-based construction~\citep{forman2003bochner,sreejith2016forman}, and associated Forman geometric flows have previously been studied on weighted complex networks~\citep{weber2017formanflow}. Curvature has also been applied to molecular and binding-affinity problems. Persistent Forman--Ricci descriptors have been constructed from three-dimensional protein--ligand complexes~\citep{wee2021fprc}, while CurvAGN incorporates curvature-based graph representations for protein--ligand affinity prediction~\citep{wu2023curvagn}. Dynamic curvature has subsequently been incorporated into graph-learning architectures. CurvFlow-Transformer uses curvature-flow-based masked attention for molecular property prediction~\citep{chen2024graphcurvatureflow}, whereas Graph Neural Ricci Flow couples time-varying curvature with continuous node-feature evolution on general attributed graphs~\citep{chen2025graphneuralricciflow}. Recent theoretical work has further studied weighted Forman and Lin--Lu--Yau Ricci flows and their convergence properties on graphs~\citep{bai2026weightedricciflow}. These studies establish important precedents for curvature evolution, but differ from CurvFlow-DTA in the prediction task, graph representation, or manner in which curvature flow is incorporated into the learning architecture.

\runinhead{Curvature-aware DTA and our position.}
Within DTA, Ricci-GraphDTA is the most closely related prior model to CurvFlow-DTA. It incorporates static Forman curvature into the molecular graph encoder, while representing the target with a sequence-based BiLSTM--attention module~\citep{zheng2026riccigraphdta}. CurvFlow-DTA instead uses weighted Forman curvature-flow trajectories on both the molecular graph and the protein residue contact graph. These trajectories are computed independently for individual entities, while a pair-conditioned selector determines the flow horizon read by each branch for a particular drug--target pair. The resulting framework therefore combines dual-graph curvature evolution with pair-conditioned flow-scale selection, distinguishing it from both static curvature-aware DTA models and curvature-flow methods developed for single-entity graph learning tasks.

\section{Methods}\label{sec:methods}

\begin{figure*}[t]
\centering
\includegraphics[width=\textwidth]{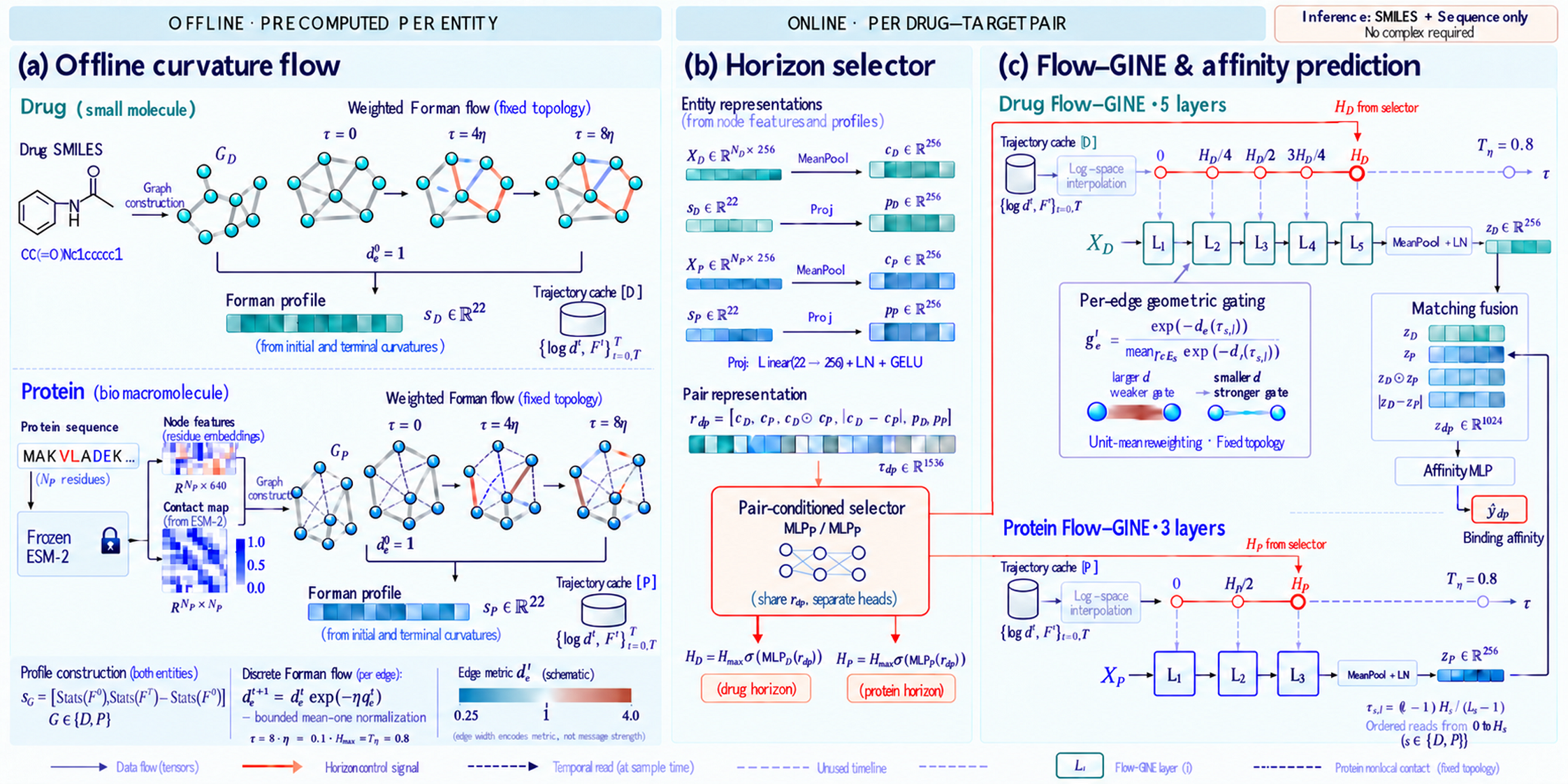}
\caption{Overview of CurvFlow-DTA. \textbf{(a)}~For each drug and target, weighted
Forman curvature flow evolves the edge metrics of the molecular graph and the
ESM-2--derived residue contact graph at fixed topology. The resulting trajectory
and Forman profile are precomputed once per entity and reused across all pairs.
\textbf{(b)}~For a given drug--target pair, a pair-conditioned selector reads the
pooled features and Forman profiles of both entities and predicts two horizons
$H_D,H_P$ that set how far each branch reads its trajectory. \textbf{(c)}~Drug and
protein Flow-GINE encoders read the selected curvature states through a per-edge
geometric gate, and their pooled representations are fused to predict affinity
$\hat{y}$. Inference requires only a SMILES string and a protein sequence. Molecule and graphs are schematic. This explanatory workflow diagram was prepared with assistance from OpenAI Codex (GPT-5.6 Sol), used to help create the workflow illustration.}
\label{fig:architecture}
\end{figure*}

\subsection{Problem formulation and model overview}

Given a drug $d$, a target protein $p$, and a measured binding affinity $y_{dp}$, the DTA task is to learn a predictor
$\hat y_{dp}=f_{\theta}(d,p)$.
CurvFlow-DTA represents the drug as a molecular graph $G_D$ and the target as a residue contact graph $G_P$, and separates graph-geometric preprocessing from pair-specific prediction.

Previous curvature-aware DTA models employ static curvature representations and encode each drug identically, regardless of its paired target. By contrast, for each unique drug or protein entity, CurvFlow-DTA uses an entity-specific stage to precompute a label-independent weighted Forman curvature-flow trajectory of edge metrics over a fixed graph topology. This trajectory depends only on the corresponding entity graph and is cached once for reuse across all drug--target pairs containing that entity. For a particular pair $(d,p)$, a pair-conditioned selector then determines separate flow horizons for the drug and protein branches. The same entity can therefore be read at different points along its precomputed trajectory when paired with different partners, without recomputing the curvature flow.

CurvFlow-DTA comprises four main components (Fig.~\ref{fig:architecture}): (i) construction of the molecular and protein residue graphs; (ii) offline weighted Forman curvature flow that generates an ordered edge-metric trajectory for each entity; (iii) pair-conditioned flow-horizon selection that determines the horizon read by each branch; and (iv) drug and protein Flow-GINE encoders that incorporate the selected edge metrics into message passing, followed by representation fusion and affinity prediction. The following subsections describe these components in detail.

\subsection{Molecular and protein graph construction}

\runinhead{Molecular graph.}
For each drug, we construct a molecular graph $G_D=(V_D,E_D)$ from its SMILES representation using RDKit. For multi-fragment inputs, the largest organic connected component is retained. Nodes correspond to heavy atoms and edges to covalent bonds, with standard RDKit atom and bond attributes used as node and edge features (Supplementary~\ref{SI-sup:features}). The Forman curvature used in this work is defined on the undirected graph, so each covalent bond is treated as a single undirected edge during curvature-flow computation and is duplicated into two directed edges only for message passing. Self-loops are not included in the curvature calculation and are handled through the GINE self-term.

All molecular edges are initialized with unit metric, $d_e^{D,0}=1$. Thus, the initial edge metric encodes only which atoms are bonded and is not derived from hand-set bond lengths or a 3D conformer, while variations in the initial Forman curvature arise solely from the molecular graph topology. Subsequent curvature-flow updates evolve these edge metrics while leaving the molecular connectivity fixed.

\runinhead{Protein residue contact graph.} 
Each target protein of length $N$ is represented as a residue graph
$G_P=(V_P,E_P)$, with one node for each residue. The initial node representation concatenates a frozen ESM-2 residue embedding (\path{esm2_t30_150M_UR50D})~\citep{lin2023esm2}, an amino-acid-type embedding, and a sinusoidal positional encoding. The concatenated features are projected to the hidden dimension by a linear layer followed by LayerNorm.

The protein graph contains backbone and long-range contact edges. Backbone edges $e_{(i,i+1)}$ connect consecutive residues and ensure graph connectivity. Long-range edges are selected from symmetrized ESM-2 contact predictions. We consider residue pairs with sequence separation $|i-j|\ge 6$ to exclude near-neighbour pairs dominated by local sequence connectivity. Candidate contacts are ranked by their predicted contact scores in descending order, and we take the global top-$K$ ($K=N$) and cap each residue's long-range degree at $8$.

The contact scores are used only to determine graph connectivity and are not used as initial edge metrics. Each retained edge is assigned the unit metric $d_e^{P,0}=1$, while its edge feature indicates whether it is a backbone or long-range contact edge. Details of contact selection and the processing of sequences exceeding the ESM-2 context length are provided in Supplementary~\ref{SI-sup:features}.

\subsection{Weighted Forman curvature and offline flow}\label{sec:flow}

For an edge $e=(u,v)$ with metric $d_e^t>0$ and unit node weights, we use the one-dimensional weighted Forman curvature~\citep{forman2003bochner,sreejith2016forman},
\begin{equation*}
F_e^t=d_e^t\!\left[
\frac{1}{d_e^t}+\frac{1}{d_e^t}
-\!\!\sum_{\substack{f\ni u\\ f\ne e}}\!\frac{1}{\sqrt{d_e^t d_f^t}}
-\!\!\sum_{\substack{g\ni v\\ g\ne e}}\!\frac{1}{\sqrt{d_e^t d_g^t}}
\right].
\end{equation*}
With the unit initial metric, this reduces to
$F_e^0=4-\deg(u)-\deg(v)$.
Thus, the initial curvature is determined by local graph topology, with lower values assigned to edges whose endpoints have larger combined degree. After each metric update, the curvature is recomputed from the evolved edge metrics in $O(|E|)$, so subsequent curvature states depend on both the fixed graph topology and the evolving relative edge geometry.

Following the general idea of evolving edge metrics under Forman curvature~\citep{weber2017formanflow}, we use a stabilized and graph-normalized update to generate an ordered trajectory of edge metrics. At step $t$, the length-weighted mean curvature is
\begin{equation*}
\bar F^t=
\frac{\sum_{e\in E}d_e^tF_e^t}
     {\sum_{e\in E}d_e^t}.
\end{equation*}
A fixed per-graph curvature scale is estimated from the initial state as
\begin{equation*}
\xi_G=
\operatorname{RMS}_{e\in E}
\left(F_e^0-\bar F^0\right)+\epsilon,
\end{equation*}
where $\epsilon>0$ prevents division by zero when the initial curvature variation is zero. The metric is then updated by
\begin{equation*}
\begin{aligned}
q_e^t
=\tanh\!\left(
\frac{F_e^t-\bar F^t}{\xi_G}
\right),
\widetilde d_e^{t+1}
=d_e^t\exp(-\eta q_e^t).
\end{aligned}
\end{equation*}
The transformed curvature deviation $q_e^t\in(-1,1)$ sets the update direction. The $\tanh$ transformation bounds the magnitude of the log-metric update, and the fixed scale $\xi_G$ reduces sensitivity to differences in curvature dispersion across graphs. The multiplicative update is equivalent to an additive step in $\log d_e^t$ and preserves metric positivity before normalization. Edges with below-average curvature are lengthened, whereas edges with above-average curvature are shortened, producing a sequence of curvature-guided edge metrics for downstream message passing.

The provisional metric $\widetilde d^{t+1}$ is mapped to the bounded, unit-mean constraint set
\begin{equation*}
\mathcal C=
\left\{
d:\ell\le d_e\le u,\;
\frac{1}{|E|}\sum_{e\in E}d_e=1
\right\}
\end{equation*}
through
\begin{equation*}
\begin{aligned}
d_e^{t+1}
&=
\left[\Pi_{\mathcal C}
\left(\widetilde d^{t+1}\right)\right]_e
=
\operatorname{clip}
\left(
c_t\,\widetilde d_e^{t+1},
\ell,u
\right),\\
1&=\frac{1}{|E|}\sum_{e\in E} d_e^{t+1}.
\end{aligned}
\end{equation*}
Here $\Pi_{\mathcal C}$ applies a single global scale $c_t>0$ and then clips. $c_t$ is the unique value for which the clipped metrics have mean one, found by one-dimensional bisection. This bounded mean normalization keeps the flow numerically stable, preventing individual metrics from collapsing toward zero or diverging while removing variation in the global metric scale.

We use $\ell=0.25$, $u=4$, $T=8$ flow steps, and a step size $\eta=0.1$, giving a maximum flow horizon $H_{\max}=T\eta=0.8$. For each unique entity, we cache the states $\{\log d^t\}_{t=0}^{T}$ and the corresponding curvatures $\{F^t\}_{t=0}^{T}$. Log-metric states are stored because both the multiplicative flow update and the trajectory interpolation described in Section~\ref{sec:horizon} are naturally expressed in log space.

We additionally compute a $22$-dimensional Forman response profile,
\begin{equation*}
s_G=
\left[
\operatorname{Stats}(F^0),\,
\operatorname{Stats}(F^T)-\operatorname{Stats}(F^0)
\right],
\end{equation*}
which summarizes the initial curvature distribution and its change over the cached trajectory. The profile is used only by the pair-conditioned horizon selector. Its component statistics and numerical conventions are given in Supplementary~\ref{SI-sup:features}, and the profile standardization parameters are fit using the training data only.

\subsection{Pair-conditioned flow horizons}\label{sec:horizon}

Let $X_D$ and $X_P$ denote the drug and protein node representations after projection to the common hidden dimension and before Flow-GINE message passing. Coarse entity representations are obtained by mean pooling,
\begin{equation*}
c_D=\operatorname{MeanPool}(X_D),\qquad
c_P=\operatorname{MeanPool}(X_P).
\end{equation*}
A lightweight pair-conditioned selector combines these representations with the Forman response profiles,
\begin{equation*}
r_{dp}=
[\,c_D,\,
c_P,\,
c_D\odot c_P,\,
|c_D-c_P|,\,
\operatorname{Proj}(s_D),\,
\operatorname{Proj}(s_P)\,],
\end{equation*}
and predicts separate flow horizons,
\begin{equation*}
\begin{aligned}
H_D&=
H_{\max}\,
\sigma\!\bigl(\mathrm{MLP}_D(r_{dp})\bigr),\\
H_P&=
H_{\max}\,
\sigma\!\bigl(\mathrm{MLP}_P(r_{dp})\bigr).
\end{aligned}
\end{equation*}
The drug and protein profiles use two separate projections, each a $\mathrm{Linear}(22\to 256)$ followed by LayerNorm and GELU. For a particular pair, its horizons $H_D$ and $H_P$ are bounded by $H_{\max}$ and determine how far along the corresponding cached trajectory each branch reads. The summary profiles $s_D$ and $s_P$ are used only by the horizon selector. They are not provided to the Flow-GINE encoders or the affinity decoder. The encoders receive curvature-flow information through the selected edge metrics.

For branch $s\in\{D,P\}$ with $L_s$ Flow-GINE layers, the selected horizon is distributed uniformly across layers,
\begin{equation*}
\tau_{s,l}
=
\frac{l-1}{L_s-1}H_s,
\qquad
l=1,\ldots,L_s.
\end{equation*}
Thus, the first layer always uses the initial metric ($\tau_{s,1}=0$), the last layer uses the selected horizon ($\tau_{s,L_s}=H_s$), and the intermediate layers read the trajectory at ordered flow times between these endpoints.

The offline trajectory contains the cached states $\{d^t\}_{t=0}^{T}$, where $t$ is the discrete flow-step index and $d^t$ corresponds to flow time $t\eta$. Hence the cached trajectory covers the interval $[0,T\eta]=[0,H_{\max}]$. For an arbitrary read
time $\tau\in[0,H_{\max}]$, let
\begin{equation*}
j(\tau)=
\min\!\left(
\left\lfloor\frac{\tau}{\eta}\right\rfloor,
T-1
\right),
\qquad
\delta(\tau)=
\frac{\tau}{\eta}-j(\tau).
\end{equation*}
The metric at $\tau$ is obtained by linear interpolation in log space,
\begin{equation*}
\log d_e(\tau)
=
[1-\delta(\tau)]\log d_e^{j(\tau)}
+
\delta(\tau)\log d_e^{j(\tau)+1}.
\end{equation*}
This definition gives $d_e(0)=d_e^0$ and $d_e(H_{\max})=d_e^T$, explicitly handling the terminal cached state. The resulting layer schedule preserves the order of the flow trajectory. Successive Flow-GINE layers read nondecreasing flow times from the initial state to the pair-selected horizon rather than treating the cached states as an unordered collection of edge features.

\subsection{Flow-GINE encoders and affinity prediction}

For branch $s\in\{D,P\}$, the interpolated metric at layer $l$ is converted into a positive scalar gate for each undirected edge,
\begin{equation*}
g_e^{(l)}
=
\frac{
\exp\!\bigl(-d_e(\tau_{s,l})\bigr)
}{
\operatorname{mean}_{f\in E_s}
\exp\!\bigl(-d_f(\tau_{s,l})\bigr)
}.
\end{equation*}
This fixed exponential mapping assigns smaller gates to edges with larger flow metrics and larger gates to edges with smaller metrics. The normalization enforces unit mean over the undirected edge set, so the flow changes the relative weighting of edges without introducing a graph-dependent shift in the mean gate value. For message passing, the same gate is assigned to the two directed copies of each undirected edge.

The gated messages are aggregated within a residual GINE update~\citep{xu2019gin,hu2020strategies},
\begin{equation*}
\begin{aligned}
m_v^{(l)}
&=
\sum_{u\in\mathcal N(v)}
g_{uv}^{(l)}
\phi_l\!\left(h_u^{(l-1)},e_{uv}\right),\\
h_v^{(l)}
&=
\operatorname{LayerNorm}\!\left[
h_v^{(l-1)}\right.\\
&\quad+\left.
\operatorname{MLP}_l
\left(
(1+\epsilon_l)h_v^{(l-1)}+m_v^{(l)}
\right)
\right].
\end{aligned}
\end{equation*}
Here $h_v^{(l)}$ denotes the representation of node $v$, $e_{uv}$ the corresponding edge feature, and $\phi_l$ the edge-aware message transformation used by the GINE layer. The geometric gate therefore modulates the contribution of each existing edge according to the flow metric read at the current layer. It changes message weights without altering graph topology.

The drug and protein encoders use the same message-passing formulation but have independent parameters. The drug branch contains $5$ layers and the protein branch $3$ layers. Both use hidden dimension $256$ and dropout $0.1$. After the final Flow-GINE layer, we use global mean pooling on each branch to produce one representation for each entity. We use mean rather than sum pooling so that the pooled representation does not scale directly with the number of atoms or residues. The pooled representations are normalized with LayerNorm and combined as
\begin{equation*}
z_{dp}
=
[\,z_D,\,
z_P,\,
z_D\odot z_P,\,
|z_D-z_P|\,].
\end{equation*}
A two-layer MLP maps the fused representation $z_{dp}$ to the predicted affinity $\hat y_{dp}$.

\section{Results}

\subsection{Experimental setup}\label{sec:exp_setup}

\runinhead{Datasets and data splits.}
We evaluate CurvFlow-DTA on the Davis~\citep{davis2011comprehensive} and KIBA~\citep{tang2014making} benchmarks. Davis dissociation constants are converted to $pK_d$, whereas the integrated KIBA scores are used as the regression targets for KIBA. Dataset statistics and the complete preprocessing details are provided in Supplementary~\ref{SI-sup:data} and \ref{SI-sup:features}.

For warm evaluation, we use the five-fold split files released with DeepDTA~\citep{ozturk2018deepdta}, alongside an independent held-out test set. In our model-selection protocol, folds $0$--$3$ are used for training and fold $4$ is used for validation, including early stopping and hyperparameter selection. After model selection, the selected configuration is refit on folds $0$--$4$ and evaluated once on the held-out test set, which is not used at any stage of model selection.

For cold-start evaluation, we construct entity-disjoint partitions without using affinity. Cold-drug splits separate compounds by Bemis--Murcko scaffold~\citep{bemis1996frameworks}, cold-target splits separate proteins by clustering their sequences with MMseqs2~\citep{steinegger2017mmseqs2} at a $40\%$ identity threshold, and dual-cold splits hold out both drug and target groups from training. The split construction targets an $80{:}10{:}10$ train/validation/test ratio.

\runinhead{Training.}
The trainable components of CurvFlow-DTA, such as the selector, both encoders, and the decoder, are trained end to end with a single Huber loss ($\delta=1$) on standardized labels. The frozen ESM-2 backbone and the precomputed curvature-flow trajectories are not updated during training.

We use AdamW with a learning rate of $10^{-4}$, weight decay of $10^{-5}$, and batch size $64$. Training proceeds for at most $300$ epochs, with early stopping after $30$ epochs without improvement in validation MSE. Full hyperparameter settings are reported in Supplementary~\ref{SI-tab:hyperparams}.

\subsection{Evaluation and comparison protocol}\label{sec:evaluation_protocol}

We evaluate prediction accuracy using mean squared error (MSE), concordance index (CI), and $R_m^2$. MSE measures regression error, CI evaluates pairwise ranking consistency, and $R_m^2$ assesses agreement between measured and predicted affinities in external validation. Predictions are transformed back to the original dataset label scale before evaluation, and all reported metrics are computed on that scale.

The comparison combines literature-reported results with a controlled reproduction. Complete tables and the source-selection rationale are provided in Supplementary~\ref{SI-sup:baseline_inventory}. Because Ricci-GraphDTA~\citep{zheng2026riccigraphdta} is the most closely related curvature-aware DTA model and serves as the primary methodological baseline for CurvFlow-DTA, we reproduce it within our experimental pipeline. To ensure a controlled comparison, the reproduction uses the same data preprocessing, split files, validation-based model-selection procedure, and evaluation code as CurvFlow-DTA. In particular, the test partition is reserved exclusively for final evaluation and is not used for hyperparameter selection.

\subsection{Warm-split performance}

CurvFlow-DTA shows strong performance under the warm-split evaluation (Table~\ref{tab:warm_results}). In the controlled comparison with the reproduced Ricci-GraphDTA baseline, CurvFlow-DTA improves all three reported metrics on both datasets. On Davis, it achieves a CI of $0.902$, an MSE of $0.185$, and an $R_m^2$ of $0.711$, compared with $0.891$, $0.231$, and $0.699$, respectively, for Ricci-GraphDTA. The reduction in MSE is approximately $20\%$. On KIBA, CurvFlow-DTA reduces MSE from $0.169$ to $0.137$ (approximately $19\%$), while increasing CI from $0.868$ to $0.887$ and $R_m^2$ from $0.735$ to $0.790$. These consistent gains under a common preprocessing, model-selection, and evaluation protocol show that the improvement over the closest curvature-aware baseline persists independently of cross-study protocol differences.

The numerical comparison with the broader set of published baselines is also favorable. CurvFlow-DTA has the lowest Davis MSE ($0.185$) and the highest CI ($0.902$). Its $R_m^2$ of $0.711$ is close to, but below, GRA-DTA ($0.715$). On KIBA, CurvFlow-DTA has the lowest listed MSE ($0.137$), and the highest $R_m^2$ ($0.790$). Its CI of $0.887$ is below the best listed value of $0.891$, reported for GraphDTA (GAT--GCN) and MGPLI. It attains the best numerical value in four of the six warm metric columns, with consistent advantages in regression error. Thus, CurvFlow-DTA combines clear improvements over the protocol-matched curvature baseline with performance at or near the strongest reported results across both warm-split benchmarks.

\begin{figure}[t]
\centering
\includegraphics[width=\columnwidth]{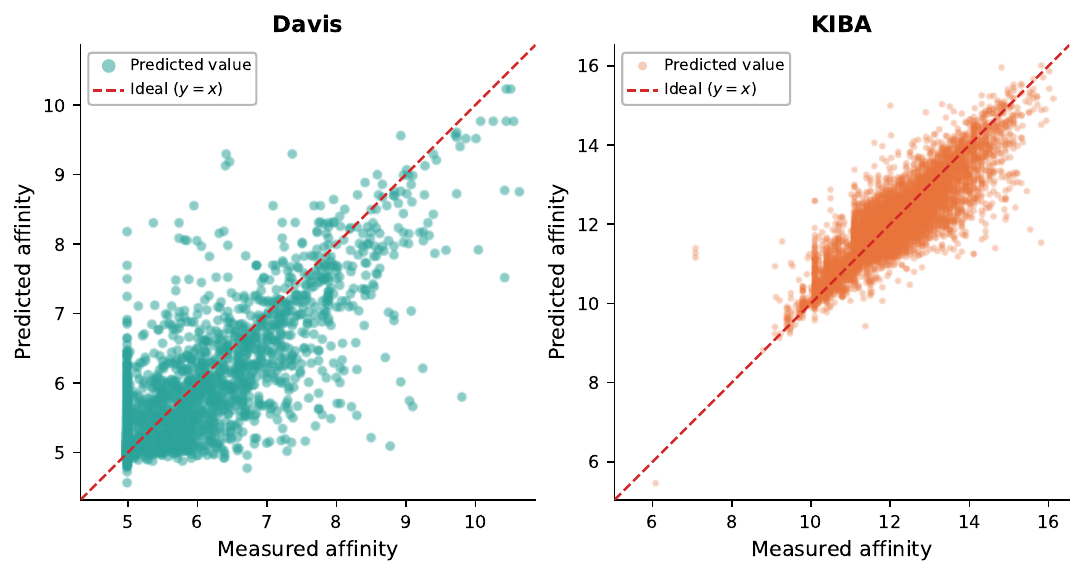}
\caption{Predicted versus measured affinities for CurvFlow-DTA on the
Davis and KIBA warm-split test sets. Each point represents one
drug--target pair, and the dashed identity line denotes perfect agreement
($y=x$).}
\label{fig:scatter}
\end{figure}

Predicted-versus-measured affinities in Fig.~\ref{fig:scatter} show clear overall correspondence between predictions and measurements on both test sets, while also revealing the residual dispersion and the characteristic concentration of Davis observations at $pK_d=5$.

\begin{table*}[t]
\centering
\small
\caption{Warm-split prediction performance on Davis and KIBA. Boldface indicates the best numerical
value among the entries listed for each metric and dataset}
\label{tab:warm_results}
\setlength{\tabcolsep}{4pt}
\begin{tabular*}{\textwidth}{@{\extracolsep{\fill}}lcccccc@{}}
\toprule
& \multicolumn{3}{c}{Davis} & \multicolumn{3}{c}{KIBA} \\
\cmidrule(lr){2-4}\cmidrule(lr){5-7}
Method & MSE $\downarrow$ & CI $\uparrow$ & $R_m^2\uparrow$ & MSE $\downarrow$ & CI $\uparrow$ & $R_m^2\uparrow$ \\
\midrule
DeepDTA~\citep{ozturk2018deepdta} & $0.261$ & $0.878$ & $0.630$ & $0.194$ & $0.863$ & $0.673$ \\
GraphDTA (GIN)~\citep{nguyen2021graphdta} & $0.229$ & $0.893$ & --- & $0.147$ & $0.882$ & --- \\
GraphDTA (GAT-GCN)~\citep{nguyen2021graphdta} & $0.245$ & $0.881$ & --- & $0.139$ & $\mathbf{0.891}$ & --- \\
FusionDTA~\citep{yuan2022fusiondta} & $0.226$ & $0.891$ & $0.686$ & $0.155$ & $0.882$ & $0.750$ \\
MGraphDTA~\citep{yang2022mgraphdta} & $0.228$ & $0.883$ & $0.679$ & $0.152$ & $0.884$ & $0.766$ \\
MGPLI~\citep{wang2022mgpli} & $0.218$ & $0.884$ & $0.620$ & $0.159$ & $\mathbf{0.891}$ & $0.753$ \\
GRA-DTA~\citep{tang2024gradta} & $0.225$ & $0.897$ & $\mathbf{0.715}$ & $0.142$ & $0.890$ & $0.784$ \\
DMFF~\citep{he2025dmff} & $0.218$ & $0.894$ & $0.702$ & $0.144$ & $0.889$ & $0.773$ \\
\midrule
Ricci-GraphDTA~\citep{zheng2026riccigraphdta} & $0.231$ & $0.891$ & $0.699$ & $0.169$ & $0.868$ & $0.735$ \\
\textbf{CurvFlow-DTA} & $\mathbf{0.185}$ & $\mathbf{0.902}$ & $0.711$ & $\mathbf{0.137}$ & $0.887$ & $\mathbf{0.790}$ \\
\bottomrule
\end{tabular*}
\end{table*}

\subsection{Cold-start performance}

Cold-start evaluation provides a more stringent test of prediction on entities not observed during model training. Our cold-drug, cold-target, and dual-cold settings hold out drug scaffold groups, protein homology groups, or both. Table~\ref{tab:cold_start_results} summarizes representative literature results and the matched comparison. The complete inventory is provided in Supplementary~\ref{SI-sup:baseline_inventory}. Because $R_m^2$ is not consistently available for the literature baselines under cold-start evaluation, the comparative table focuses on MSE and CI. Shared scenario names specify which entity is held out, but do not imply identical test entities or grouping rules across publications.

\begin{table*}[t]
\centering
\small
\caption{Cold-start results for representative baselines. Boldface indicates the best numerical value among the entries listed within each dataset and scenario.}
\label{tab:cold_start_results}
\setlength{\tabcolsep}{4pt}
\begin{tabular*}{\textwidth}{@{\extracolsep{\fill}}lcccccc@{}}
\toprule
& \multicolumn{2}{c}{Cold-drug} & \multicolumn{2}{c}{Cold-target} & \multicolumn{2}{c}{Dual-cold} \\
\cmidrule(lr){2-3}\cmidrule(lr){4-5}\cmidrule(lr){6-7}
Method & MSE $\downarrow$ & CI $\uparrow$ & MSE $\downarrow$ & CI $\uparrow$ & MSE $\downarrow$ & CI $\uparrow$ \\
\midrule
\multicolumn{7}{l}{\textbf{Davis}} \\
GraphDTA & $0.557$ & $0.735$ & $0.439$ & $0.770$ & $0.656$ & $0.607$ \\
MGPLI & $0.895$ & $0.584$ & $0.457$ & $0.792$ & $0.654$ & $0.556$ \\
GRA-DTA & $0.578$ & $0.725$ & $0.376$ & $0.827$ & $0.560$ & $0.690$ \\
GDilatedDTA & $0.500$ & $0.761$ & $0.382$ & $0.810$ & $0.718$ & $0.633$ \\
FusionDTA & $0.581$ & $0.737$ & $0.364$ & $0.826$ & $0.876$ & $0.645$ \\
MGraphDTA & $0.563$ & $0.729$ & $0.359$ & $0.813$ & $0.874$ & $0.636$ \\
DMFF & $0.548$ & $0.742$ & $0.330$ & $\mathbf{0.840}$ & $0.759$ & $0.655$ \\
\cmidrule(lr){1-7}
Ricci-GraphDTA & $0.554$ & $0.734$ & $0.413$ & $0.796$ & $0.616$ & $0.696$ \\
\textbf{CurvFlow-DTA} & $\mathbf{0.475}$ & $\mathbf{0.782}$ & $\mathbf{0.300}$ & $0.833$ & $\mathbf{0.492}$ & $\mathbf{0.723}$ \\
\midrule
\multicolumn{7}{l}{\textbf{KIBA}} \\
GraphDTA & $0.349$ & $0.761$ & $0.519$ & $0.651$ & $0.473$ & $0.647$ \\
MGPLI & $0.500$ & $0.725$ & $0.531$ & $0.673$ & $0.640$ & $0.611$ \\
GRA-DTA & $\mathbf{0.337}$ & $\mathbf{0.765}$ & $0.435$ & $0.692$ & $\mathbf{0.441}$ & $0.657$ \\
FusionDTA & $0.429$ & $0.748$ & $0.439$ & $0.685$ & $0.587$ & $0.641$ \\
MGraphDTA & $0.425$ & $0.746$ & $0.435$ & $0.674$ & $0.590$ & $0.626$ \\
DMFF & $0.408$ & $0.753$ & $0.410$ & $\mathbf{0.748}$ & $0.567$ & $0.667$ \\
\cmidrule(lr){1-7}
Ricci-GraphDTA & $0.406$ & $0.726$ & $0.462$ & $0.710$ & $0.534$ & $0.590$ \\
\textbf{CurvFlow-DTA} & $0.347$ & $0.748$ & $\mathbf{0.383}$ & $0.746$ & $0.445$ & $\mathbf{0.674}$ \\
\bottomrule
\end{tabular*}
\end{table*}

On Davis, CurvFlow-DTA has the lowest listed MSE in all three cold-start settings. It also has the highest listed CI under cold-drug ($0.782$) and dual-cold ($0.723$) evaluation. Under cold-target evaluation, its CI of $0.833$ is second to DMFF ($0.840$). Relative to the Ricci-GraphDTA values on Davis, CurvFlow-DTA reduces MSE by $14.3\%$, $27.4\%$, and $20.1\%$ under cold-drug, cold-target, and dual-cold evaluation, respectively, while improving CI in all three settings.

On KIBA, CurvFlow-DTA has the lowest listed cold-target MSE ($0.383$), while its CI of $0.746$ is second to DMFF ($0.748$). Under cold-drug evaluation, its MSE of $0.347$ is second to GRA-DTA ($0.337$), and its CI of $0.748$ is below the best listed value ($0.765$, GRA-DTA). Under dual-cold evaluation, CurvFlow-DTA has the highest listed CI ($0.674$), and its MSE of $0.445$ is close to GRA-DTA ($0.441$). It therefore has the lowest or second-lowest MSE across all six cold-start settings, including four numerical minima, while ranking performance varies with the scenario. In the matched KIBA comparison, CurvFlow-DTA improves over Ricci-GraphDTA in all three cold-start settings, reducing MSE from $0.406$ to $0.347$, from $0.462$ to $0.383$, and from $0.534$ to $0.445$. These correspond to reductions of $14.5\%$, $17.1\%$, and $16.7\%$, respectively. CI increases from $0.726$ to $0.748$, from $0.710$ to $0.746$, and from $0.590$ to $0.674$.

The matched cold-start results show that our gains over the closest curvature-aware baseline hold for unseen drugs and targets on both benchmarks. Comparisons with other published results further show consistently low regression error and competitive ranking performance.

\subsection{Ablation analysis}

To examine the contribution of the geometric design choices, we compare CurvFlow-DTA with a graded set of controls.
\emph{No curvature} sets $g_e^{(l)}\equiv1$ for every edge and layer, retaining the same dual-graph encoders and affinity decoder while removing curvature-based edge gating.
\emph{Static Forman} keeps a curvature-derived gate but never advances the flow. The per-edge gate is computed once from the initial Forman curvature $F^0$ and held fixed across all layers, so the model still sees curvature geometry but only as a single static snapshot rather than an evolving trajectory. This isolates the value of evolving the geometry from the value of using curvature and provides a static curvature-gating comparator.
\emph{Fixed flow} removes the pair-conditioned horizon selector and fixes $H_D=H_P=H_{\max}$ for every drug--target pair. Each branch nevertheless retains the ordered curvature-flow trajectory and reads it according to the same pair-independent layer-wise flow-time schedule.
\emph{Entity-only horizon} restores adaptive horizon selection but determines the horizon from each entity independently, without information from its paired partner.
\emph{Metric-flow shuffle} preserves the cached metric trajectories and their per-step distributions but randomly reassigns trajectories among edges, breaking the correspondence between an edge and its own flow evolution.
Finally, \emph{Drug-flow only} and \emph{Protein-flow only} enable curvature flow and apply flow-based gating to only one branch while retaining uniform gating ($g_e\equiv 1$) on the other.
Tables~\ref{tab:ablation_results} and~\ref{tab:ablation_results_ci} report MSE and CI across warm and all three cold-start settings.

\begin{table*}[t]
\centering
\caption{Ablation analysis of CurvFlow-DTA in terms of MSE
($\downarrow$) across warm, cold-drug, cold-target, and dual-cold
evaluation on Davis and KIBA. The controls modify the
curvature representation, trajectory readout, or graph branch. Boldface indicates the
best value in each column.}
\label{tab:ablation_results}
\small
\setlength{\tabcolsep}{4pt}
\renewcommand{\arraystretch}{1.15}

\begin{threeparttable}
\begin{tabular}{lcccccccc}
\toprule
\multirow{2}{*}{Model variant}
& \multicolumn{4}{c}{Davis}
& \multicolumn{4}{c}{KIBA} \\
\cmidrule(lr){2-5}\cmidrule(lr){6-9}
& Warm & Cold-drug & Cold-target & Dual-cold
& Warm & Cold-drug & Cold-target & Dual-cold \\
\midrule
No curvature        & 0.231 & 0.513 & 0.401 & 0.542 & 0.151 & 0.363 & 0.406 & 0.470 \\
Static Forman      & 0.221 & 0.501 & 0.344 & 0.523 & 0.148 & 0.358 & 0.399 & 0.465 \\
Fixed flow         & 0.204 & 0.487 & 0.323 & 0.507 & 0.143 & 0.352 & 0.392 & 0.454 \\
Entity-only horizon & 0.194 & 0.481 & 0.311 & 0.498 & 0.140 & 0.349 & 0.387 & 0.449 \\
Metric-flow shuffle  & 0.226 & 0.526 & 0.360 & 0.560 & 0.153 & 0.374 & 0.403 & 0.483 \\
Drug-flow only     & 0.218 & 0.497 & 0.440 & 0.553 & 0.149 & 0.357 & 0.411 & 0.476 \\
Protein-flow only  & 0.226 & 0.532 & 0.453 & 0.548 & 0.150 & 0.377 & 0.399 & 0.474 \\
Full CurvFlow-DTA
& \textbf{0.185} & \textbf{0.475} & \textbf{0.300} & \textbf{0.492}
& \textbf{0.137} & \textbf{0.347} & \textbf{0.383} & \textbf{0.445} \\
\bottomrule
\end{tabular}
\end{threeparttable}
\end{table*}

\begin{table*}[t]
\centering
\caption{Ablation analysis of CurvFlow-DTA in terms of CI ($\uparrow$) across warm, cold-drug, cold-target, and dual-cold
evaluation on Davis and KIBA. Boldface indicates the best value in
each column.}
\label{tab:ablation_results_ci}
\small
\setlength{\tabcolsep}{4pt}
\renewcommand{\arraystretch}{1.15}

\begin{threeparttable}
\begin{tabular}{lcccccccc}
\toprule
\multirow{2}{*}{Model variant}
& \multicolumn{4}{c}{Davis}
& \multicolumn{4}{c}{KIBA} \\
\cmidrule(lr){2-5}\cmidrule(lr){6-9}
& Warm & Cold-drug & Cold-target & Dual-cold
& Warm & Cold-drug & Cold-target & Dual-cold \\
\midrule
No curvature        & 0.889 & 0.764 & 0.807 & 0.700 & 0.883 & 0.739 & 0.726 & 0.662 \\
Static Forman      & 0.895 & 0.771 & 0.817 & 0.708 & 0.884 & 0.742 & 0.733 & 0.667 \\
Fixed flow         & 0.899 & 0.777 & 0.828 & 0.716 & 0.885 & 0.745 & 0.740 & 0.671 \\
Entity-only horizon & 0.901 & 0.780 & 0.831 & 0.720 & 0.886 & 0.746 & 0.743 & 0.672 \\
Metric-flow shuffle  & 0.899 & 0.756 & 0.824 & 0.688 & 0.882 & 0.728 & 0.732 & 0.648 \\
Drug-flow only     & 0.899 & 0.771 & 0.801 & 0.694 & 0.884 & 0.743 & 0.732 & 0.655 \\
Protein-flow only  & 0.897 & 0.748 & 0.775 & 0.697 & 0.883 & 0.724 & 0.738 & 0.658 \\
Full CurvFlow-DTA
& \textbf{0.902} & \textbf{0.782} & \textbf{0.833} & \textbf{0.723}
& \textbf{0.887} & \textbf{0.748} & \textbf{0.746} & \textbf{0.674} \\
\bottomrule
\end{tabular}
\end{threeparttable}
\end{table*}

The graded controls show a consistent monotonic ordering. Across both datasets and all four evaluation settings, performance improves along \emph{No curvature} $\rightarrow$ \emph{Static Forman} $\rightarrow$ \emph{Fixed flow} $\rightarrow$ \emph{Entity-only horizon} $\rightarrow$ the full model, for both MSE and CI. This pattern supports the incremental value of curvature-informed gating, evolved edge metrics, adaptive trajectory readout, and pair-conditioned horizon selection. The magnitude of the improvement varies across settings. The largest MSE gain over \emph{No curvature} occurs for Davis cold-target prediction, where MSE decreases from $0.401$ to $0.300$, a relative reduction of approximately $25\%$. For CI, the improvement over no curvature is larger in every cold-start setting than on the corresponding warm split for both datasets.

The three intermediate controls, \emph{Static Forman}, \emph{Fixed flow} and \emph{Entity-only horizon}, further show that it is the evolution of curvature, not merely its presence, that contributes to the performance improvement. \emph{Static Forman} improves over no curvature in every reported cell, showing that curvature-informed gating is already useful without flow evolution. Replacing static curvature with an ordered flow trajectory under a fixed, pair-independent horizon yields a further improvement throughout. Allowing an entity-dependent horizon improves performance again, and conditioning that horizon on the complete drug--target pair gives the best result in every cell. For example, Davis cold-target MSE decreases successively from $0.401$ with \emph{No curvature} to $0.344$ with \emph{Static Forman}. Introducing the ordered flow trajectory with a fixed pair-independent horizon further reduces MSE to $0.323$. Allowing the horizon to adapt to each entity lowers it to $0.311$, and conditioning the horizon on the complete drug--target pair gives the best value of $0.300$.

The metric-flow shuffle provides a complementary control on whether the observed improvement can be explained by added edge-metric variations. Although this variant preserves the per-step distribution of cached metrics, it performs below the full model in every MSE and CI column and falls below \emph{No curvature} in several settings. This indicates that the distribution of flow-derived edge metrics alone is insufficient. Their correspondence with the graph edges is important for predictive performance.

The single-branch controls also favor applying flow-based gating on both entity graphs. The full model outperforms both \emph{Drug-flow only} and \emph{Protein-flow only} in every reported setting. Notably, for Davis cold-target prediction, the two single-branch variants yield MSE values of $0.440$ and $0.453$, respectively, compared with $0.401$ for \emph{No curvature} and $0.300$ for the full model. Thus, applying flow to only one side is not sufficient to reproduce the gain of the dual-graph formulation and can underperform \emph{No curvature} in some settings.

Because the \emph{No curvature} control retains the same downstream encoder and decoder capacity, the consistent advantage of the full model also argues against parameter count alone as an explanation for the observed gains. Taken together, these ablations support the combined contribution of curvature-guided edge gating, flow evolution, pair-conditioned horizon selection, and flow-based encoding on both drug and protein graphs.

\subsection{Case study: protein attribution and crystallographic binding pockets}

If CurvFlow-DTA transfers to unseen targets by reading genuine protein geometry, the residues its protein branch relies on most should fall inside the true binding pocket, even on a target held out of training. We test this directly on two cold-target complexes.

We attribute predictions on the protein contact graph with an edge-level counterfactual. For each undirected edge we restore its flow metric to the initial state $t{=}0$, and record the resulting change in predicted affinity. A residue's importance is the summed importance of its incident edges, averaged as a percentile rank across the five trained seeds. The ground-truth pocket is defined structurally as the set of protein residues within $4$\,\AA{} of any ligand heavy atom in the co-crystal. Because high-degree residues accumulate incident edges and could dominate any edge-based score, we benchmark each residue-importance ranking against two null models: the random pocket proportion and the residue degree itself.

We present two representative cold-target test pairs, both kinases with an exact high-resolution co-crystal and a five-seed mean prediction error within our preset acceptance range ($<1.5$). Both are held out under the cold-target split. The first is MET (UniProt P08581), a receptor tyrosine kinase and established oncology target, bound to a small-molecule inhibitor (PDB 3U6H, ligand 03X, $2.0$\,\AA, 21 pocket residues). PDB 3U6H is a wild-type human c-MET construct, so its sequence matches the model input. The second is cyclin-dependent kinase 2 (CDK2, UniProt P24941), a central cell-cycle regulator, bound to olomoucine (PDB 1W0X, $2.20$\,\AA, 12 pocket residues).

\begin{figure*}[t]
\centering
\includegraphics[width=\textwidth]{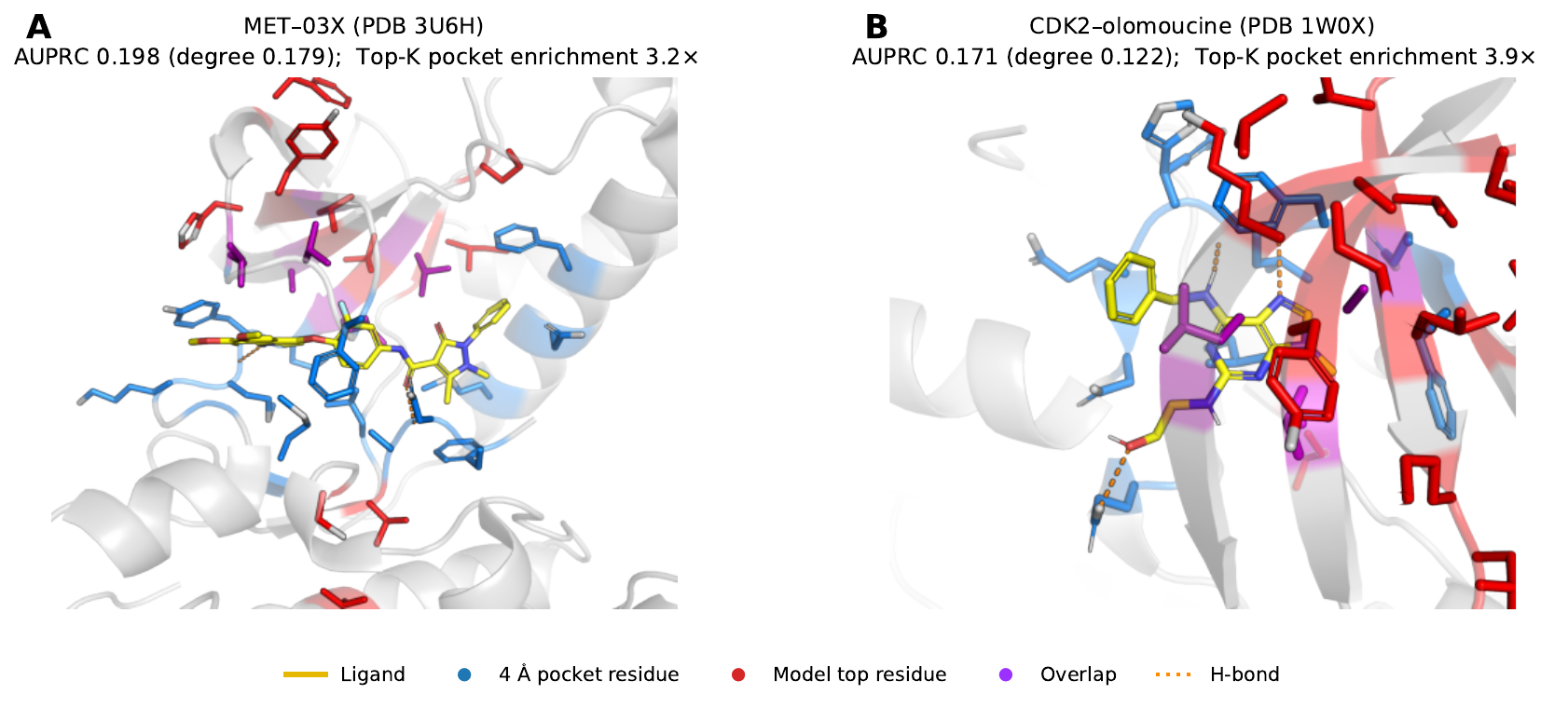}
\caption{Protein-graph attribution on two held-out cold-target co-crystals:
\textbf{(A)}~MET--03X (PDB 3U6H) and \textbf{(B)}~CDK2--olomoucine
(PDB 1W0X), shown as PyMOL binding-pocket close-ups. The inhibitor is drawn
as yellow sticks, the $4$\,\AA{} crystallographic pocket residues in blue, the
model's top-$20$ attributed residues in red, residues in both in purple, and
ligand--protein hydrogen bonds as orange dashes. Each panel is annotated with
the pocket-recovery AUPRC (versus the residue-degree baseline) and the top-$K$
enrichment over chance. Both crystal structures are wild-type constructs
matching the model's input sequences.}
\label{fig:casestudy}
\end{figure*}

For this MET complex, residue importance attains an AUPRC of $0.198$, above both the random ($0.075$) and residue-degree ($0.179$) baselines. It recovers $3$ and $5$ pocket residues among the top-$10$ and top-$20$ attributed residues, a $3.2\times$ enrichment over chance among its top-$K$ residues ($K$ the pocket size). The five-seed prediction error on this pair is small ($0.024$). The recovered residues are hydrophobic liners of the ATP cleft (Ile1084, Val1092, Ala1108, Val1155, and Leu1157), rather than surface residues. For CDK2–olomoucine the pattern holds. It yields an AUPRC of $0.171$ (random $0.043$, degree $0.122$), representing a $40.2\%$ improvement over the degree baseline, and a $3.9\times$ enrichment. Top-ranked residues such as Ile10, Val18, and Ala31 line the ATP pocket. In both complexes the highest-importance residues concentrate around the ATP-binding cleft rather than spreading over the surface or tracking node degree (Fig.~\ref{fig:casestudy}).

The attribution exceeds the residue-degree baseline, so the ranking is not merely a preference for well-connected residues. The model reads only sequence at inference and never sees the bound complex, yet its most important residues fall in the crystallographic pocket. These two selected examples show enrichment of attributed residues in crystallographic binding pockets, supporting the structural relevance of the protein-branch attribution in these cases.

\section{Discussion}

Two patterns in these results help explain the behavior of CurvFlow-DTA. First, replacing a single static curvature representation with an ordered curvature-flow trajectory provides the encoder with a range of geometric scales rather than one fixed scale. The pair-conditioned selector further allows the same entity to be read at different flow horizons when paired with different partners. This introduces curvature-guided message-reweighting patterns that can be read at a pair-dependent horizon.

Second, applying the same curvature-aware formulation to the protein residue contact graph allows the model to retain graph structure rather than representing the target only through a sequence-level summary. ESM-2 supplies frozen residue representations and sequence-derived contact information, while curvature flow acts on the resulting contact graph through the evolving edge metrics. Importantly, the ablation controls retain the same protein features and contact-graph construction while altering or removing the flow-based edge reweighting. Their consistent gap from the full model therefore shows that the observed benefit cannot be explained by the protein language model or contact representation alone, and supports an additional contribution from curvature-guided processing of the protein graph.

The selected baseline compilation places CurvFlow-DTA among strong DTA predictors. It has the lowest listed MSE on both warm benchmarks and in four of the six cold settings, with second-lowest MSE in the other two. The controlled evidence is the improvement over Ricci-GraphDTA in all eight dataset--setting combinations, supported by ablations that retain the same protein features and graph construction. For the two structural case studies, protein-branch importance corresponds to crystallographic pocket residues without using the bound complex during prediction.

Several limitations define the scope of the present study. The evaluation is restricted to two kinase-dominated benchmarks. Literature baselines use different dataset versions, partitions, and model-selection procedures, so their numerical ranks do not establish superiority under a common test set. The attribution analysis concerns selected structural examples rather than a population-wide localization assessment. These boundaries motivate broader benchmarks and controlled comparisons using shared entity-group partitions.

\section{Conclusion}

We introduced CurvFlow-DTA, a drug--target affinity model that replaces static discrete Ricci curvature with a weighted Forman curvature flow and applies it on both molecular and protein residue-contact graphs. A dual-branch Flow-GINE reads the resulting curvature-flow trajectories, while a pair-conditioned selector determines the flow horizon used by each branch for a given drug--target pair.

Across Davis and KIBA, CurvFlow-DTA consistently improves on protocol-matched Ricci-GraphDTA under warm, cold-drug, cold-target, and dual-cold evaluation. The literature-reported comparison further shows low regression error across the two benchmarks, subject to source-specific evaluation protocols. Ablations show that each design choice contributes: removing curvature gating, freezing the flow, fixing the horizon, or restricting flow to a single graph each degrades accuracy. These results support dual-graph curvature flow as an effective component of DTA prediction, including prediction for previously unseen entities.


\section*{CRediT authorship contribution statement}
Jicheng Ma (Conceptualization, Methodology, Software, Formal analysis,
Investigation, Writing -- original draft), Yunyan Yang (Investigation,
Validation, Writing -- review \& editing), Juan Zhao (Data curation,
Validation, Writing -- review \& editing), and Liang Zhao (Conceptualization,
Supervision, Funding acquisition, Writing -- review \& editing).

\section*{Funding}
This work was supported by the National Natural Science Foundation of China [No. 12271039]. The funder had no role in the study design, data processing, preparation of the manuscript, or decision to submit the article for publication.

\putbib[references]

\end{bibunit}
\clearpage
\arxivsupplementsetup
\begin{bibunit}[abbrvnat-jmgm]
\arxivbibnamespace{SI}

\begin{frontmatter}
\title{Supplementary information for ``CurvFlow-DTA: dual-graph discrete Ricci curvature flow for drug--target affinity prediction''}
\author{Jicheng Ma}
\author{Yunyan Yang}
\author{Juan Zhao}
\author{Liang Zhao\corref{cor1}}
\cortext[cor1]{Corresponding author.}
\begin{abstract}
This document provides supplementary details for the main article, including dataset statistics and split construction, molecular and protein graph features, evaluation metrics, extended results, the offline weighted Forman curvature-flow algorithm, training convergence, implementation settings, and the complete hyperparameter settings.
\end{abstract}
\end{frontmatter}

\section{Datasets and split construction}\label{SI-sup:data}

\mypar{Benchmarks.}
We use the Davis~\citep{davis2011comprehensive} and KIBA~\citep{tang2014making} benchmarks, following the label processing convention from DeepDTA~\citep{ozturk2018deepdta}. Davis provides dissociation constants in nanomolar units, which are converted to \(pK_d\) values via
\begin{equation*}
pK_d
=
-\log_{10}\!({K_d}/{10^9})
=
9-\log_{10}\!K_d.
\end{equation*}
The KIBA dataset provides KIBA scores that integrate $K_d$, $K_i$, and $\mathrm{IC}_{50}$ measurements. Both datasets can be organized as drug-target affinity matrices. Only experimentally measured entries are used, and each labelled drug-target pair represents one regression target.
Table~\ref{SI-tab:datastats} summarizes the basic statistics of the two datasets.

\begin{figure*}[t]
\centering
\includegraphics[width=0.92\textwidth]{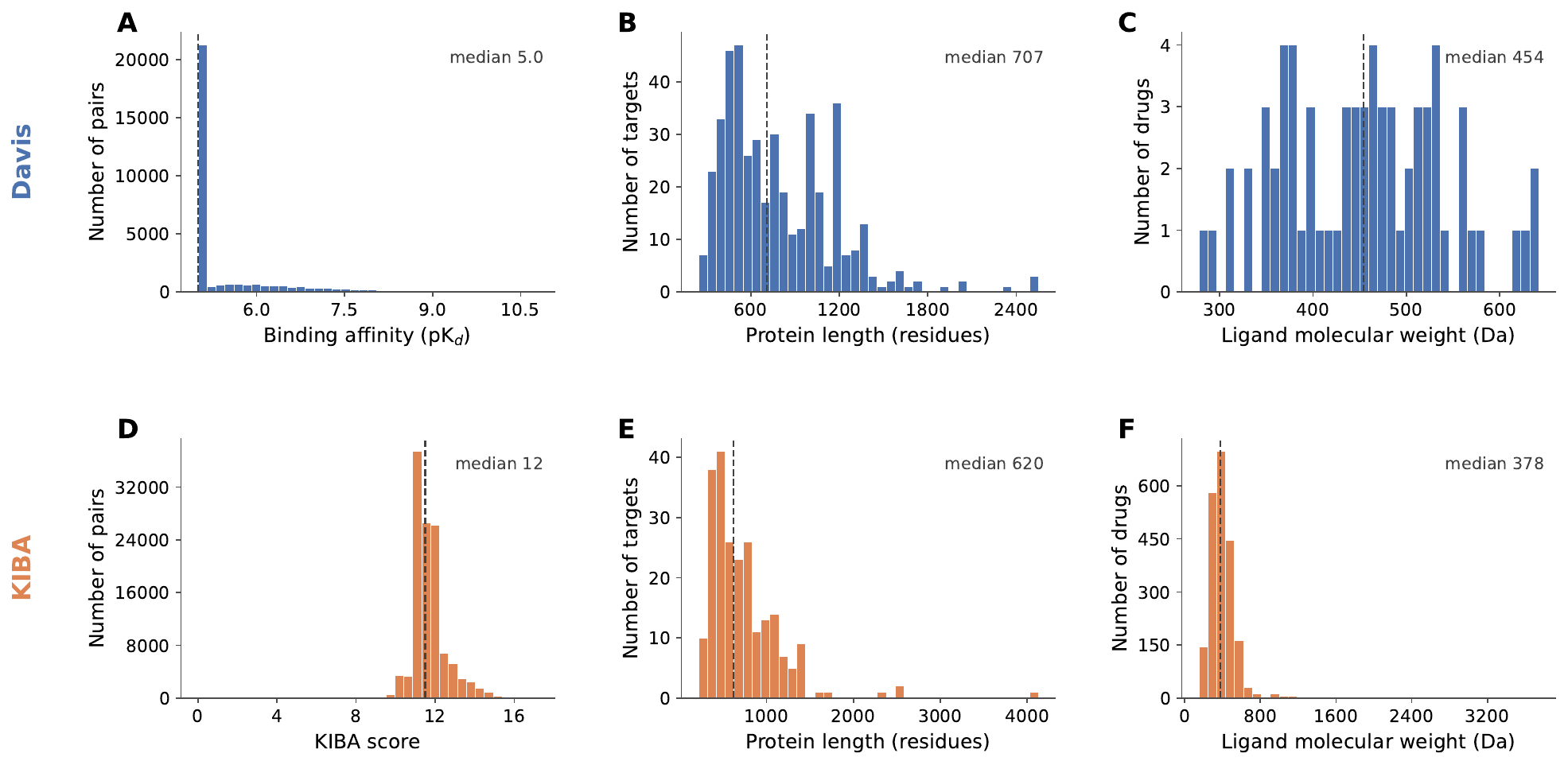}
\caption{Distributions of benchmark labels (\textbf{A}, \textbf{D}),
protein sequence lengths (\textbf{B}, \textbf{E}), and molecular
weights (\textbf{C}, \textbf{F}) for Davis (top row) and KIBA
(bottom row). Dashed vertical lines indicate the medians. 
Molecular weights are computed using RDKit.}
\label{SI-fig:datastats}
\end{figure*}

The concentration of Davis labels at $pK_d=5$ (see A in Fig.~\ref{SI-fig:datastats}) includes weak-binding or undetected interactions encoded at the benchmark limit. These entries should not all be interpreted as precise measurements of $K_d=10\,000\,\mathrm{nM}$~\citep{ozturk2018deepdta}.

\begin{table}[ht]
\centering
\small
\caption{Benchmark statistics. Counts are for the full measured matrix. The
warm-split train/test sizes follow the released DeepDTA setting-1 folds used unchanged in
this work.}
\label{SI-tab:datastats}
\begin{tabular}{lrr}
\toprule
 & Davis & KIBA \\
\midrule
Drugs                       & 68     & 2\,111 \\
Targets                     & 442    & 229 \\
Measured pairs              & 30\,056 & 118\,254 \\
Warm train pairs (folds 0--4) & 25\,046 & 98\,545 \\
Warm test pairs             & 5\,010  & 19\,709 \\
\bottomrule
\end{tabular}
\end{table}

\mypar{Warm split.} For warm evaluation, we use the train/test split files released with
DeepDTA~\citep{ozturk2018deepdta}. The released non-test portion is organized into five folds. In our model-selection protocol, folds $0$--$3$ are used for training and fold $4$ is used for validation, including early stopping and hyperparameter selection. After model selection, the selected configuration is refit on folds $0$--$4$ and evaluated once on the held-out test set, which is not used at any stage of model selection.

\mypar{Cold-start splits.} The cold-start splits are entity-disjoint and are constructed without using affinity values (label-blind), so that no measurement influences the partition. Cold‑drug isolates drugs by Bemis‑Murcko scaffold~\citep{bemis1996frameworks} using RDKit. For cold‑target, protein sequences are clustered with MMseqs2~\citep{steinegger2017mmseqs2} at a $40\%$ sequence-identity threshold, which groups the $442$ Davis targets into $222$ clusters and the $229$ KIBA targets into $127$ clusters. Each cluster is assigned in full to one of the training, validation, or test sets, so that no test target shares more than $40\%$ identity with any training target. Dual‑cold applies both constraints and holds out unseen drugs and targets simultaneously. Each split targets an $80{:}10{:}10$ train/validation/test ratio with respect to labelled pairs. After construction, we perform disjointness checks corresponding to each evaluation setting. We verify scaffold separation for cold-drug, sequence-cluster separation for cold-target, and both forms of separation for dual-cold. The partitions are fixed across all five training seeds.

\section{Molecular and protein graph features}\label{SI-sup:features}

\mypar{Molecular graph.} Drug SMILES are parsed using RDKit~\citep{rdkitsoftware}. For multi-fragment inputs, the largest organic connected component is retained. If two components tie, we apply deterministic tie‑breaking. We first select the component with more heavy atoms, followed by the one with a larger total atom count. If ties persist, we choose the component whose canonical SMILES appears earlier in alphabetical order. The molecular graph contains one node per heavy atom and one undirected edge per covalent bond. Atom attributes comprise element type (one-hot over the observed element set), degree, explicit valence, formal charge, aromaticity, hybridization state, ring membership, chirality, and total hydrogen count. Together, these attributes assemble into a $53$‑dimensional atom feature vector. Bond attributes comprise bond type (single/double/triple/aromatic), conjugation, ring membership, and stereochemistry, collectively forming a $12$‑dimensional bond feature vector. These attributes are used as the molecular node and edge features, respectively.

Curvature-flow computation uses the unique undirected edge set. At each flow step, each undirected edge has one metric and one curvature value. For message passing, an undirected edge is represented by two directed copies that share the same flow metric and geometric gate. Self-loops are excluded from curvature-flow computation and handled by the GINE self-term.

\mypar{Protein contact graph.} For a protein of length $N$, the residue graph is $G_P=(V_P,E_P)$ with $V_P=\{1,\ldots,N\}$. Its edge set contains backbone edges and long-range contact edges. Backbone edges connect consecutive residues,
\begin{equation*}
E_{\mathrm{bb}}
=
\bigl\{\{i,i+1\}:i=1,\ldots,N-1\bigr\},
\end{equation*}
which ensures connectivity of the residue graph. Long-range contact candidates are obtained from symmetrized ESM-2 contact predictions. Symmetrization is implemented by taking the element‑wise average of the raw contact matrix and its transpose. Only residue pairs satisfying $|i-j|\ge6$ are considered, excluding pairs dominated by local sequence connectivity. Then we take the global top-$K$ ($K=N$) and cap each residue's long-range degree at $8$. Candidate contacts are ranked by their
predicted contact scores in descending order (ties broken by residue index). We then traverse the ranked candidates and add an edge only when both of its endpoints still have long-range degree below $8$, otherwise we skip it and move on. We stop as soon as $K$ edges have been added. If the degree cap prevents us from ever reaching $K$, we simply keep the fewer edges we could add rather than relaxing the cap. Backbone edges are always kept and are not subject to this limit. The ESM-2 contact probabilities are used only to determine which long-range edges are retained. 

For a protein of length $N$, the input representation of residue $i$ is the concatenation of three components. The first component is a frozen ESM-2 residue representation $\mathbf e_i\in\mathbb R^{640}$ extracted from the $30$th representation layer of \path{esm2_t30_150M_UR50D}~\citep{lin2023esm2} with the terminal start/end tokens removed. The second component is a trainable amino-acid-type embedding. Let $t_i\in\{1,\ldots,21\}$ denote the residue type, with $20$ indices corresponding to the standard amino acids and one additional index used for unknown or nonstandard residues. The embedding is
\begin{equation*}
\mathbf a_i
=
\mathbf W_{\mathrm{aa}}[t_i],
\qquad
\mathbf W_{\mathrm{aa}}\in\mathbb R^{21\times d_a},
\end{equation*}
where $d_a=256$ is the amino-acid embedding dimension. The third component is a fixed sinusoidal positional encoding $\mathbf p_i\in\mathbb R^{d_p}$, where $d_p=256$. For $k=0,\ldots,d_p/2-1$,
\begin{equation*}
\begin{aligned}
p_{i,2k}
&=
\sin\!\left(
\frac{i}
     {10000^{2k/d_p}}
\right),\\
p_{i,2k+1}
&=
\cos\!\left(
\frac{i}
     {10000^{2k/d_p}}
\right).
\end{aligned}
\end{equation*}

The three components are concatenated and projected to the common hidden dimension,
\begin{equation*}
\mathbf h_i^{(0)}
=
\operatorname{LayerNorm}
\left(
\mathbf W_P
[\,\mathbf e_i \Vert \mathbf a_i \Vert \mathbf p_i\,]
+
\mathbf b_P
\right),
\end{equation*}
with $\mathbf W_P\in\mathbb R^{256\times(640+d_a+d_p)}$ and $\mathbf b_P\in\mathbb R^{256}$. Here $\mathbf W_{\mathrm{aa}}$, $\mathbf W_P$, and $\mathbf b_P$ are trainable and ESM-2 is frozen. The resulting $\mathbf h_i^{(0)}$ is the initial residue node representation supplied to the protein Flow-GINE encoder.

\mypar{Long-sequence ESM-2 processing.}
For proteins longer than $1022$ residues, ESM-2 inputs are divided into overlapping residue windows of length $1022$ with an overlap of $256$ residues. ESM-2 representations are extracted independently for each window. Precisely, windows start at residues $1,1+s,1+2s,\ldots$ with stride $s=1022-256=766$, and the loop stops once a window reaches residue $N$. The final window is simply truncated at $N$ and may therefore be shorter than $1022$.

A residue pair only receives a contact score if the two residues fall inside the same window at least once. A pair that never co-occurs in any window is simply excluded from the candidate set and it can never become an edge. In the implementation their contact scores are held at a sentinel value below every real score and masked out by a co-occurrence mask. When a pair does co-occur in several overlapping windows, we average its scores over those windows, and likewise average each residue's embedding over the windows covering it. The full-length matrix is assembled by writing each window's block back into its own position and dividing by the number of windows that covered each entry.

For proteins with length $N>1022$, we first symmetrize the contact‑score matrix after the stitching step. We then apply the $|i-j|\ge6$ filter, the global top-$K$ ranking, and the per-residue degree cap during long-range edge selection. The positional encoding uses the global residue index $i\in\{1,\ldots,N\}$ of the full sequence. Windowing only affects how the frozen ESM-2 embeddings and contact scores are produced and never resets the position index. All of this reduces to the single-window case for proteins short enough to fit in one window.

\mypar{Forman response profile.}
For the curvature-flow trajectory, define
\begin{equation*}
\begin{aligned}
\operatorname{Stats}(F)
&=
\bigl[\,\mu_F,\sigma_F,\gamma_F,\kappa_F,
\min F,\max F, \\
&\qquad Q_{10},Q_{25},Q_{50},Q_{75},Q_{90}\,\bigr],
\end{aligned}
\end{equation*}
which collects the mean, standard deviation, skewness, excess kurtosis, minimum, maximum, and the $10/25/50/75/90$ percentiles of the edge-curvature distribution. When the standard deviation $\sigma_F\leq 10^{-8}$, both skewness and excess kurtosis are set to $0$. This occurs only for graphs with constant edge curvature. The response profile is
\begin{equation*}
s_G
=
[\,\operatorname{Stats}(F^0),\,
\operatorname{Stats}(F^T)-\operatorname{Stats}(F^0)\,]
\in\mathbb R^{22}.
\end{equation*}
It contains the initial curvature summary and the net change in these summary statistics between the initial and terminal cached states. The profile is supplied only to the horizon selector and profile standardization statistics are fit on the training set only.

\section{Evaluation protocol and metrics}\label{SI-sup:eval}

\mypar{Reported statistics.} Every model trained in our own pipeline (CurvFlow-DTA and Ricci-GraphDTA) is run over five random seeds and reported as mean $\pm$ sample standard deviation. Literature values are quoted from their source publications. Label standardization and profile scaling are fit on training data only and applied unchanged to validation and test.

\mypar{Metric definitions.}
For $n$ evaluated drug--target pairs with labels $y_i$ and predictions $\hat y_i$, let $\bar y$ and $\overline{\hat y}$ be the corresponding sample means. The Pearson correlation $r$ is
\begin{equation*}
r=
\frac{\sum_i(y_i-\bar y)(\hat y_i-\overline{\hat y})}
{\sqrt{\sum_i(y_i-\bar y)^2
       \sum_i(\hat y_i-\overline{\hat y})^2}}.
\end{equation*}
MSE, CI and $R_m^2$ are defined as
\begin{equation*}
\mathrm{MSE}=
\frac1n\sum_{i=1}^{n}(y_i-\hat y_i)^2,
\end{equation*}
\begin{equation*}
\mathrm{CI}
=
\frac{1}{|\mathcal Q|}
\sum_{(i,j)\in\mathcal Q}
\left[
\mathbf 1(\hat y_i>\hat y_j)
+\frac12\mathbf 1(\hat y_i=\hat y_j)
\right],
\end{equation*}
\begin{equation*}
R_m^2=r^2\left(1-\sqrt{\,r^2-r_0^2\,}\right),
\end{equation*}
where $\mathcal Q=\{(i,j):y_i>y_j\}$, and $r_0^2$ is the coefficient of determination of the regression through the origin,
\begin{equation*}
k_0=
\frac{\sum_i y_i\hat y_i}{\sum_i\hat y_i^2},
\quad
r_0^2=
1-
\frac{\sum_i(y_i-k_0\hat y_i)^2}
     {\sum_i(y_i-\bar y)^2}.
\end{equation*}
A larger $R_m^2$ (with $r^2\ge r_0^2$) indicates better agreement on held-out data.

The main text reports CI, MSE, and $R_m^2$. Lower MSE and higher CI and $R_m^2$ indicate better performance. In addition, we report the complete set of metrics in Tables~\ref{SI-tab:ext_warm} and~\ref{SI-tab:ext_cold}, covering results from both benchmarks, evaluated on warm and cold‑start splits. Reported metrics include MSE, RMSE, CI, \(R_m^2\), and Pearson $r$, which are presented as mean \(\pm\) sample standard deviation over five random seeds.

\begin{table*}[ht]
\centering
\small
\caption{Extended warm-split results for CurvFlow-DTA.
Values are mean $\pm$ sample standard deviation over five
training seeds.}
\label{SI-tab:ext_warm}
\begin{tabular*}{\textwidth}{@{\extracolsep{\fill}}lccccc@{}}
\toprule
Dataset & MSE $\downarrow$ & RMSE $\downarrow$ & CI $\uparrow$
        & $R_m^2 \uparrow$ & Pearson $\uparrow$ \\
\midrule
Davis & $0.185 \pm 0.016$ & $0.430 \pm 0.019$ & $0.902 \pm 0.004$ & $0.711 \pm 0.019$ & $0.861 \pm 0.004$ \\
KIBA  & $0.137 \pm 0.002$ & $0.370 \pm 0.002$ & $0.887 \pm 0.002$ & $0.790 \pm 0.007$ & $0.893 \pm 0.001$ \\
\bottomrule
\end{tabular*}
\end{table*}

\begin{table*}[ht]
\centering
\small
\caption{Extended cold-start results for CurvFlow-DTA.
Values are mean $\pm$ sample standard deviation over five
training seeds.}
\label{SI-tab:ext_cold}
\begin{tabular*}{\textwidth}{@{\extracolsep{\fill}}llccccc@{}}
\toprule
Dataset &  & MSE $\downarrow$ & RMSE $\downarrow$ & CI $\uparrow$
        & $R_m^2 \uparrow$ & Pearson $\uparrow$ \\
\midrule
\multirow{3}{*}{Davis}
 & Cold-drug   & $0.475 \pm 0.018$ & $0.689 \pm 0.013$ & $0.782 \pm 0.010$ & $0.275 \pm 0.009$ & $0.546 \pm 0.015$ \\
 & Cold-target & $0.300 \pm 0.029$ & $0.547 \pm 0.027$ & $0.833 \pm 0.010$ & $0.329 \pm 0.037$ & $0.638 \pm 0.014$ \\
 & Dual-cold   & $0.492 \pm 0.033$ & $0.701 \pm 0.023$ & $0.723 \pm 0.047$ & $0.071 \pm 0.018$ & $0.305 \pm 0.052$ \\
\midrule
\multirow{3}{*}{KIBA}
 & Cold-drug   & $0.347 \pm 0.032$ & $0.588 \pm 0.027$ & $0.748 \pm 0.013$ & $0.449 \pm 0.032$ & $0.682 \pm 0.026$ \\
 & Cold-target & $0.383 \pm 0.015$ & $0.619 \pm 0.012$ & $0.746 \pm 0.006$ & $0.454 \pm 0.022$ & $0.706 \pm 0.011$ \\
 & Dual-cold   & $0.445 \pm 0.059$ & $0.666 \pm 0.044$ & $0.674 \pm 0.022$ & $0.182 \pm 0.067$ & $0.431 \pm 0.142$ \\
\bottomrule
\end{tabular*}
\end{table*}

\mypar{Ricci-GraphDTA reproduction.}
Ricci-GraphDTA~\citep{zheng2026riccigraphdta} is the most closely related geometry-aware DTA baseline to CurvFlow-DTA and therefore serves as our primary controlled comparator. To ensure that differences in performance are not confounded by differences in data handling or model-selection procedures, we reproduced Ricci-GraphDTA using the same preprocessing, data splits, validation protocol, and evaluation code as CurvFlow-DTA. In particular, the released Ricci-GraphDTA implementation evaluates performance on the test set after each training epoch and uses the resulting test-set performance to retain the best-performing model checkpoint. This differs from the evaluation principle adopted throughout our study, in which the test set is strictly reserved for final evaluation and is never used for hyperparameter tuning, early stopping, checkpoint selection, or any other model-selection decision. We therefore modified only the model-selection procedure so that checkpoint selection and early stopping were based exclusively on the validation set. The Ricci-GraphDTA architecture and method-specific hyperparameters are kept, and only the shared splits, validation-only selection, five-seed protocol, and common evaluation code are substituted. We reproduced Ricci-GraphDTA for the warm and Davis cold-start settings and additionally evaluated it on the KIBA cold-start splits, which were not reported in the original study.

\section{Offline weighted Forman curvature flow}\label{SI-sup:flow}

For each drug and protein graph, curvature flow is computed on the edge set with initial edge metrics $d_e^0=1$. Algorithm~\ref{SI-alg:offline-flow} gives the offline procedure that produces the cached curvature-flow trajectory. It depends only on the entity graph and never on affinity labels or on the drug--target pairing, so it is computed once per entity and reused across all pairs that involve that entity.

\begin{figure*}[t]
\centering
\small
\setlength{\fboxsep}{8pt}
\refstepcounter{algorithm}\label{SI-alg:offline-flow}
\fbox{%
\begin{minipage}{0.92\textwidth}
\textbf{Algorithm \thealgorithm.} Offline weighted Forman curvature flow (per entity)\\[2pt]
\hrule
\vspace{4pt}
\textbf{Input:} An entity graph $G=(V,E)$ with unit edge metric $d_e^{0}=1$, steps $T$, step size $\eta$, bounds $[\ell,u]=[0.25,4]$ and safeguard $\epsilon=10^{-8}$.\\
\textbf{Output:} cached states $\{\log d^{t}\}_{t=0}^{T}$, curvatures
$\{F^{t}\}_{t=0}^{T}$, and response profile $s_G$.
\vspace{4pt}
\hrule
\vspace{4pt}
\begin{tabular}{@{}r@{\quad}l@{}}
1: & Compute $F_e^{0}$ for all $e$ \quad (weighted Forman curvature) and calculate
$\bar F^0$\\[3pt]
2: & $\xi_G \leftarrow \operatorname{RMS}_e\!\bigl(F_e^{0}-\bar F^{0}\bigr)+\epsilon$ \quad \emph{// fixed per-graph scale} \\[3pt]
3: & \textbf{for} $t = 0,1,\dots,T-1$ \textbf{do} \\[3pt]
4: & \quad $\bar F^{t} \leftarrow \bigl(\textstyle\sum_e d_e^{t} F_e^{t}\bigr)/\bigl(\sum_e d_e^{t}\bigr)$ \quad \emph{// length-weighted mean} \\[3pt]
5: & \quad $q_e^{t} \leftarrow \tanh\!\bigl((F_e^{t}-\bar F^{t})/\xi_G\bigr)$ \quad \emph{// signed update direction} \\[3pt]
6: & \quad $\widetilde d_e^{t+1} \leftarrow d_e^{t}\exp(-\eta\,q_e^{t})$ \quad \emph{// multiplicative (log-space) step} \\[3pt]
7: & \quad $d^{t+1} \leftarrow \Pi_{\mathcal C}\bigl(\widetilde d^{t+1}\bigr)$ \quad \emph{// renormalize to unit mean in $[\ell,u]$} \\[3pt]
8: & \quad recompute $F_e^{t+1}$ for all $e$ \\[3pt]
9: & \textbf{end for} \\[3pt]
10: & $s_G \leftarrow [\operatorname{Stats}(F^{0}),\,\operatorname{Stats}(F^{T})-\operatorname{Stats}(F^{0})]$ \\[3pt]
11: & \textbf{return} $\{\log d^{t}\}$, $\{F^{t}\}$, $s_G$ \\
\end{tabular}
\end{minipage}}
\end{figure*}

The per-step curvature recomputation is $O(|E|)$, so a full trajectory is $O(T|E|)$ and, being cached per unique entity, adds no repeated cost across the many pairs that share a drug or a target.

\section{Training convergence}\label{SI-sup:convergence}

\begin{figure*}[t]
\centering
\includegraphics[width=0.92\textwidth]{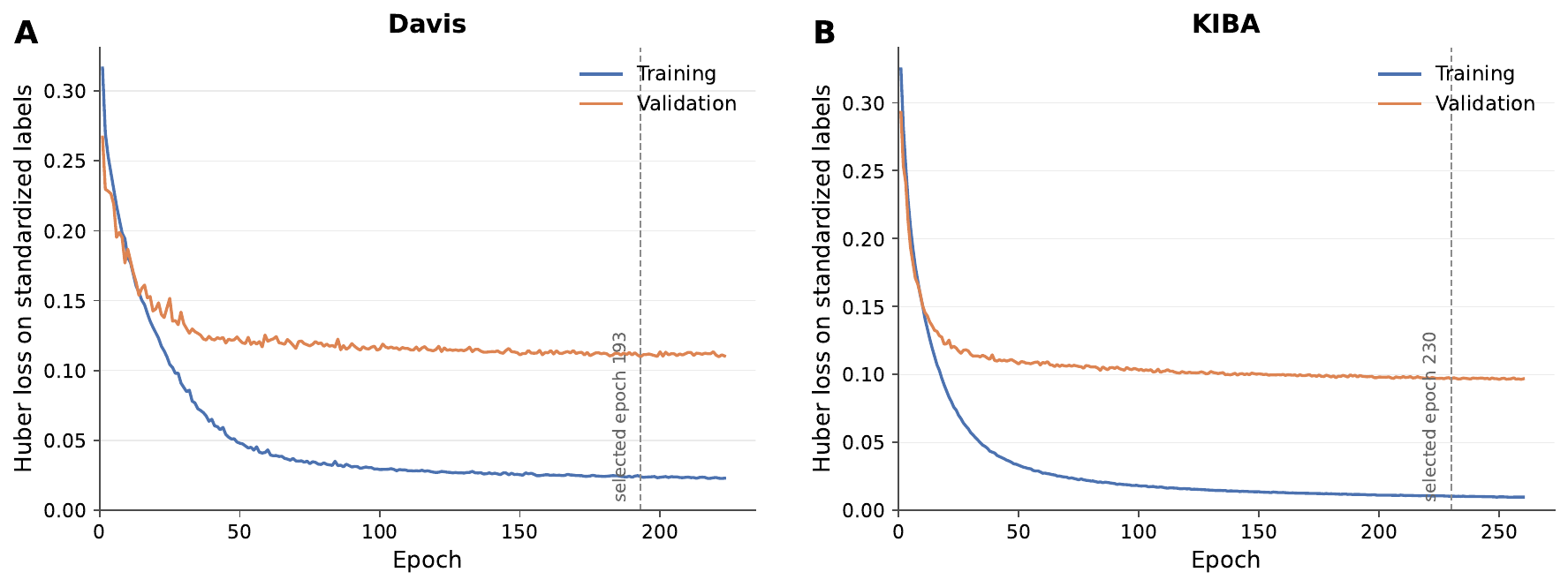}
\caption{Training and validation Huber loss versus epoch during warm-split model
selection for CurvFlow-DTA on \textbf{(A)} Davis and \textbf{(B)} KIBA (seed 0).
The dashed line marks the epoch selected by early stopping.}
\label{SI-fig:losscurve}
\end{figure*}

We plot the training and validation loss over epochs for a representative seed (seed 0) on the warm split of each benchmark (Fig.~\ref{SI-fig:losscurve}). On both datasets the loss drops quickly and then levels off, and the validation loss stays flat rather than rising, indicating stable training up to the early-stopping epoch.

\section{Hyperparameters and implementation details}\label{SI-sup:hyper}

\mypar{Hyperparameters.}
Table~\ref{SI-tab:hyperparams} lists the hyperparameters used for the reported CurvFlow-DTA runs. Further implementation details are documented below.

\begin{table}[t]
\centering
\small
\caption{Main hyperparameters of CurvFlow-DTA.}
\label{SI-tab:hyperparams}
\begin{tabularx}{\linewidth}{@{}Xl@{}}
\toprule
Component & Setting \\
\midrule
Frozen ESM-2 backbone & \path{esm2_t30_150M_UR50D} \\
Hidden dimension & 256 \\
Drug / protein encoder layers & 5 / 3 \\
Dropout & 0.1 \\
Flow steps $T$ & 8 \\
Flow step size $\eta$ & 0.1 \\
Maximum horizon $T\eta$ & 0.8 \\
Edge-metric bounds & $[0.25,4]$ \\
Safeguard $\epsilon$ & $10^{-8}$ \\
Long-range contact budget & Up to $N$ \\
Maximum long-range degree & 8 \\
Minimum sequence separation & 6 residues \\
ESM-2 window / overlap & 1022 / 256 residues \\
Huber loss threshold  & $\delta=1$ \\
Optimizer & AdamW \\
Learning rate & $10^{-4}$ \\
Weight decay & $10^{-5}$ \\
Batch size & 64 \\
Maximum training epochs & 300 \\
Early-stopping patience & 30 epochs \\
Training seeds & 5 \\
\bottomrule
\end{tabularx}
\end{table}

Implementation details are as follows. Each Forman profile is projected by its own $\mathrm{Linear}(22\to256)\to\mathrm{LayerNorm}\to\mathrm{GELU}$ (the drug and protein projections do not share weights). The selector forms the pair feature $[\,c_D,c_P,c_D\odot c_P,|c_D-c_P|,\operatorname{Proj}(s_D),\operatorname{Proj}(s_P)\,]\in\mathbb R^{1536}$ (four $256$-dim entity terms plus the two $256$-dim profile projections). Two independent horizon heads, each $\mathrm{Linear}(1536\to256)\to\mathrm{GELU}\to\mathrm{Dropout}(0.1)\to\mathrm{Linear}(256\to1)$, produce $H_D,H_P=H_{\max}\,\sigma(\cdot)$ with $H_{\max}=0.8$. All layers use default PyTorch initialization.

$\phi_l$ concatenates the node and edge features $[\,h_u\,\Vert\,e_{uv}\,]$ and passes them through a two-layer MLP, $\mathrm{Linear}(256+\dim(e_{uv})\to256)$ followed by $\mathrm{GELU}$ and $\mathrm{Linear}(256\to256)$. The GINE self-term coefficient $\epsilon_l$ is a per-layer learnable scalar initialized to $0$. It is unrelated to the curvature-flow safeguard $\epsilon=10^{-8}$.

\section{Details of ablation analysis}
\label{SI-sup:ablations}

\emph{Static Forman} builds its gate from the initial curvature $F_e^0$ rather than from the unit initial metric $d_e^0=1$. Each edge is assigned the fixed weight $g_e=\exp(-F_e^0)$, and normalized to unit mean over the undirected edge set of each graph. This gate remains fixed across all layers, and no flow update is performed. Therefore, the model sees a single non-uniform curvature snapshot instead of an evolving trajectory.

In the control of \emph{Metric-flow shuffle}, a single random permutation $\pi$ of the edge set is drawn per graph and applied to reassign whole cached metric trajectories among edges. Edge $e$ receives the entire trajectory $\{d^t\}$ originally belonging to edge $\pi(e)$. Every edge therefore still carries a valid trajectory, but the correspondence between an edge and its own curvature evolution is broken. Trajectories are permuted as whole sequences, not shuffled independently at each step.

The inactive branch of \emph{Drug-flow only} or \emph{Protein-flow only} uses uniform gating ($g_e\equiv1$), which is the same as \emph{No-curvature} control. Only the active branch reads its flow-based gate. The pair-conditioned selector still receives both entities' coarse features and Forman profiles and predicts the horizon for the active branch as usual.

\section{Protein attribution and structural case studies}
\label{SI-sup:attribution}

\mypar{Scope and case selection.}
The main text presents MET--03X and CDK2--olomoucine as two illustrative cold-target examples.
Ligand-bound crystal structures are used retrospectively to define structural
reference pockets, not as inputs to CurvFlow-DTA.

The available reference structures are PDB 3U6H, a wild-type human c-MET
kinase-domain complex with the inhibitor 03X at $2.0$\,\AA{}
~\citep{pdb3u6h}, and PDB 1W0X, a human CDK2 complex with
olomoucine at $2.20$\,\AA{}~\citep{pdb1w0x}. Both are wild-type constructs whose
sequence matches the model input. We deliberately avoid mutant co-crystals
(for example the D1228V structure PDB 6SDD) so that the reference pocket and the
model's input sequence correspond to the same protein.

\mypar{Edge perturbation and residue scores.}
For trained model $s$ and protein edge $e$, let $\hat y_{dp}^{(s)}$ be the unperturbed prediction. For one undirected protein edge $e$, we reset its flow-dependent metric to the initial value ($d_e=1$) for both of its directed copies, leaving every other edge's flowed metric, all node features, the selected horizons, and the drug branch unchanged. Then we re-run the forward pass to obtain $\hat y_{dp}^{(s,-e)}$ and get the resulting change in predicted affinity
\begin{equation*}
\Delta_e^{(s)}
=
|\hat y_{dp}^{(s)}
-
\hat y_{dp}^{(s,-e)}|.
\end{equation*}
A residue's importance is the sum of $\Delta_e$ over its incident edges, averaged as a percentile rank across the five seeds.

\mypar{Reference pockets and ranking metrics.}
Let $\mathcal U$ be the set of structurally evaluable residues.
The reference pocket $\mathcal P\subseteq\mathcal U$ contains
residues with at least one heavy atom within $4$\,\AA{} of
a heavy atom of the selected ligand instance.
The pocket prevalence is
\begin{equation*}
p_{\mathrm{pocket}}
=
\frac{|\mathcal P|}{|\mathcal U|}.
\end{equation*}
The residue-degree control ranks residues using their degree
in the model's protein graph, evaluated over the same
eligible residue set.

For the top-$k$ ranked residues $\mathcal T_k$, enrichment is
\begin{equation*}
\mathrm{EF}_k
=
\frac{|\mathcal T_k\cap\mathcal P|/k}
     {|\mathcal P|/|\mathcal U|}.
\end{equation*}
In the pocket-size-based comparison, $k=|\mathcal P|$.
For uniform random selection of $k$ residues from
$\mathcal U$, the expected number of pocket residues is
$k\,p_{\mathrm{pocket}}$. AUPRC is the area under the precision--recall curve of the residue importance ranking against pocket membership.

\mypar{Case-level numerical summaries.}
Table~\ref{SI-tab:case_summary} records the case-level values
reported in the main text.

\begin{table*}[t]
\centering
\small
\caption{Case-level attribution summaries.
The table reports the prevalence, degree-control AUPRC, model AUPRC, and
pocket-size-based enrichment.}
\label{SI-tab:case_summary}
\begin{tabular*}{\textwidth}
{@{\extracolsep{\fill}}lcccc@{}}
\toprule
Case & Pocket prevalence & Degree AUPRC
& Model AUPRC & Enrichment \\
\midrule
MET--03X & 0.075 & 0.179 & 0.198 & $3.2\times$ \\
CDK2--olomoucine & 0.043 & 0.122 & 0.171 & $3.9\times$ \\
\bottomrule
\end{tabular*}
\end{table*}

\section{Complete baseline comparisons and result provenance}
\label{SI-sup:baseline_inventory}

\mypar{Baseline coverage and selection.}
The comparison covers similarity-based methods (KronRLS and SimBoost), sequence-based methods (DeepDTA and FusionDTA), molecular-graph and graph--sequence models (GraphDTA, DeepGLSTM, DeepGS, MGraphDTA, MGPLI, GRA-DTA, and GDilatedDTA), and graph-fusion models (GLFA, GEFA, and DMFF). Ricci-GraphDTA is the primary controlled comparator because it is the closest curvature-aware DTA method.

Selection prioritizes relevant architectures, identifiable Davis/KIBA regression settings, traceable numerical sources, clearly defined evaluation settings, and publicly available implementations. We favor protocols close to our evaluation setting and distinguish method references from the publications supplying the results. Both CurvFlow-DTA and Ricci-GraphDTA are trained within our pipeline, where validation-based selection with a held-out test set is mandatory. For results taken directly from published literature, we document differences in evaluation protocols. We do not assume all external implementations or checkpoint selection routines have undergone independent verification. This list contains only the selected baselines and is not an exhaustive ranking of all DTA approaches.

\mypar{Method references and numerical sources.}
Table~\ref{SI-tab:baseline_sources} lists the publications providing the reported numerical values. Each method citation denotes its model architecture. The source identifier in each result table marks where the numeric results come from. We adopt values from the original paper unless noted otherwise below. We keep three decimal places for all reported numbers, rounding from original source data if necessary. Missing results are not filled in.

SimBoost~\citep{he2017simboost} and KronRLS~\citep{pahikkala2015toward} use the benchmark results reported by DeepDTA, because our warm evaluation uses DeepDTA's released split files. Both GIN and GAT--GCN variants of GraphDTA~\citep{nguyen2021graphdta} are shown on both datasets, corresponding to its best variants on Davis and KIBA, respectively. GraphDTA's original warm tables do not provide $R_m^2$, which is left unfilled.

Warm results for DeepGLSTM~\citep{mukherjee2022deepglstm} follow GDilatedDTA. All warm and cold results for FusionDTA~\citep{yuan2022fusiondta} and MGraphDTA~\citep{yang2022mgraphdta} follow DMFF~\citep{he2025dmff}. We make these substitutions because the original public training pipelines of those methods select checkpoints using the test set\footnote{DeepGLSTM: \url{https://github.com/MLlab4CS/DeepGLSTM/blob/main/training.py}. FusionDTA: repository \url{https://github.com/yuanweining/FusionDTA}, commit \texttt{df0ea5d4}, \texttt{training.py} and cold-start training scripts. MGraphDTA: repository \url{https://github.com/guaguabujianle/MGraphDTA}, commit \texttt{54386680}, \texttt{regression/train.py}.}.

All cold results for GraphDTA and MGPLI~\citep{wang2022mgpli} are taken from GRA-DTA's Table~4. All cold results for MGraphDTA~\citep{yang2022mgraphdta} are taken from DMFF's Tables~1--2, providing a consistent reporting source across the three scenarios on both benchmarks.

Ricci-GraphDTA~\citep{zheng2026riccigraphdta} is reproduced within our pipeline, including KIBA cold-start evaluation, rather than using the original test-selected results.

\mypar{Protocol compatibility.}
Our warm evaluation uses the released DeepDTA test set and refits the selected configuration on the five non-test folds. Alternative published evaluations need not use the same test pairs. In particular, DMFF reports a $7{:}1{:}2$ training/validation/test allocation for its warm comparison, whereas our final warm training/test allocation is
approximately $5{:}1$. DMFF describes entity-disjoint cold evaluation. Such entity-based splits are not established to match our scaffold and protein-homology partitions.

\mypar{Reporting conventions and availability.}
The full tables contain every baseline from our collected inventory when we have selected results for the matching dataset and evaluation setting. The results for CurvFlow-DTA and Ricci-GraphDTA in both warm and cold settings are given as mean \(\pm\) sample standard deviation across five training seeds. Our chosen sources do not provide KIBA cold-start results for GLFA, GEFA, and GDilatedDTA, so these rows are left out of the corresponding tables.

The complete warm-split results are listed in Tables~\ref{SI-tab:full_warm_davis} and~\ref{SI-tab:full_warm_kiba}. The Davis cold-start results are listed in Tables~\ref{SI-tab:full_cold_davis_0}, \ref{SI-tab:full_cold_davis_1}, and~\ref{SI-tab:full_cold_davis_2}; the KIBA cold-start results are listed in Tables~\ref{SI-tab:full_cold_kiba_0}, \ref{SI-tab:full_cold_kiba_1}, and~\ref{SI-tab:full_cold_kiba_2}.

For cold-start comparisons, since $R_m^2$ is not consistently supplied across the selected sources, we only use MSE and CI. Bold text highlights the best mean values within each full table.

\begin{table*}[p]
\centering
\small
\caption{Numerical-source identifiers used in the complete comparison.
The source of a result may differ from the original publication of the method.}
\label{SI-tab:baseline_sources}
\begin{tabularx}{\textwidth}{@{}lX@{}}
\toprule
Source & Publication and result location \\
\midrule
D18 & DeepDTA, Tables 3--6~\citep{ozturk2018deepdta} \\
G21 & GraphDTA, Tables 2 and 3~\citep{nguyen2021graphdta} \\
L20 & DeepGS, Tables 2 and 3~\citep{lin2020deepgs} \\
Z24 & GDilatedDTA, Tables 3--5~\citep{zhang2024gdilateddta} \\
W22 & MGPLI, Table 4 (full-model row)~\citep{wang2022mgpli} \\
T24 & GRA-DTA, Tables 2 and 4~\citep{tang2024gradta} \\
H25 & DMFF, Supplementary Tables 1--2 (warm); Tables 1--2 (cold)~\citep{he2025dmff} \\
N22 & GEFA, Table 1~\citep{nguyen2022gefa} \\
This study & Matched Ricci-GraphDTA reproduction and CurvFlow-DTA evaluation. \\
\bottomrule
\end{tabularx}
\end{table*}

\begin{table*}[p]
\centering
\small
\caption{Complete warm-split comparison on Davis. Numerical sources are defined in Table~\ref{SI-tab:baseline_sources}.}
\label{SI-tab:full_warm_davis}
\begin{tabular*}{\textwidth}{@{\extracolsep{\fill}}llccc@{}}
\toprule
Method & Source & MSE $\downarrow$ & CI $\uparrow$ & $R_m^2\uparrow$ \\
\midrule
SimBoost & D18 & $0.282$ & $0.872$ & $0.644$ \\
KronRLS & D18 & $0.379$ & $0.871$ & $0.407$ \\
DeepDTA & D18 & $0.261$ & $0.878$ & $0.630$ \\
GraphDTA (GIN) & G21 & $0.229$ & $0.893$ & --- \\
GraphDTA (GAT-GCN) & G21 & $0.245$ & $0.881$ & --- \\
DeepGLSTM & Z24 & $0.294$ & $0.867$ & $0.624$ \\
DeepGS & L20 & $0.252$ & $0.882$ & $0.686$ \\
FusionDTA & H25 & $0.226$ & $0.891$ & $0.686$ \\
MGraphDTA & H25 & $0.228$ & $0.883$ & $0.679$ \\
MGPLI & W22 & $0.218$ & $0.884$ & $0.620$ \\
GRA-DTA & T24 & $0.225$ & $0.897$ & $\mathbf{0.715}$ \\
GDilatedDTA & Z24 & $0.237$ & $0.885$ & $0.686$ \\
DMFF & H25 & $0.218$ & $0.894$ & $0.702$ \\
\midrule
Ricci-GraphDTA & This study & $0.231 \pm 0.019$ & $0.891 \pm 0.008$ & $0.699 \pm 0.010$ \\
\textbf{CurvFlow-DTA} & This study & $\mathbf{0.185 \pm 0.016}$ & $\mathbf{0.902 \pm 0.004}$ & $0.711 \pm 0.019$ \\
\bottomrule
\end{tabular*}
\end{table*}

\begin{table*}[p]
\centering
\small
\caption{Complete warm-split comparison on KIBA. Numerical sources are defined in Table~\ref{SI-tab:baseline_sources}.}
\label{SI-tab:full_warm_kiba}
\begin{tabular*}{\textwidth}{@{\extracolsep{\fill}}llccc@{}}
\toprule
Method & Source & MSE $\downarrow$ & CI $\uparrow$ & $R_m^2\uparrow$ \\
\midrule
SimBoost & D18 & $0.222$ & $0.836$ & $0.629$ \\
KronRLS & D18 & $0.411$ & $0.782$ & $0.342$ \\
DeepDTA & D18 & $0.194$ & $0.863$ & $0.673$ \\
GraphDTA (GIN) & G21 & $0.147$ & $0.882$ & --- \\
GraphDTA (GAT-GCN) & G21 & $0.139$ & $\mathbf{0.891}$ & --- \\
DeepGLSTM & Z24 & $0.185$ & $0.855$ & $0.705$ \\
DeepGS & L20 & $0.193$ & $0.860$ & $0.684$ \\
FusionDTA & H25 & $0.155$ & $0.882$ & $0.750$ \\
MGraphDTA & H25 & $0.152$ & $0.884$ & $0.766$ \\
MGPLI & W22 & $0.159$ & $\mathbf{0.891}$ & $0.753$ \\
GRA-DTA & T24 & $0.142$ & $0.890$ & $0.784$ \\
GDilatedDTA & Z24 & $0.156$ & $0.876$ & $0.775$ \\
DMFF & H25 & $0.144$ & $0.889$ & $0.773$ \\
\midrule
Ricci-GraphDTA & This study & $0.169 \pm 0.006$ & $0.868 \pm 0.002$ & $0.735 \pm 0.010$ \\
\textbf{CurvFlow-DTA} & This study & $\mathbf{0.137 \pm 0.002}$ & $0.887 \pm 0.002$ & $\mathbf{0.790 \pm 0.007}$ \\
\bottomrule
\end{tabular*}
\end{table*}

\begin{table*}[p]
\centering
\small
\caption{Complete Davis cold-drug comparison. Sources are defined in Table~\ref{SI-tab:baseline_sources}.}
\label{SI-tab:full_cold_davis_0}
\begin{tabular*}{\textwidth}{@{\extracolsep{\fill}}llcc@{}}
\toprule
Method & Source & MSE $\downarrow$ & CI $\uparrow$ \\
\midrule
GraphDTA & T24 & $0.557$ & $0.735$ \\
MGPLI & T24 & $0.895$ & $0.584$ \\
GRA-DTA & T24 & $0.578$ & $0.725$ \\
GDilatedDTA & Z24 & $0.500$ & $0.761$ \\
GLFA & N22 & $0.861$ & $0.670$ \\
GEFA & N22 & $0.847$ & $0.709$ \\
FusionDTA & H25 & $0.581$ & $0.737$ \\
MGraphDTA & H25 & $0.563$ & $0.729$ \\
DMFF & H25 & $0.548$ & $0.742$ \\
\midrule
Ricci-GraphDTA & This study & $0.554 \pm 0.074$ & $0.734 \pm 0.018$ \\
\textbf{CurvFlow-DTA} & This study & $\mathbf{0.475 \pm 0.018}$ & $\mathbf{0.782 \pm 0.010}$ \\
\bottomrule
\end{tabular*}
\end{table*}

\begin{table*}[p]
\centering
\small
\caption{Complete Davis cold-target comparison. Sources are defined in Table~\ref{SI-tab:baseline_sources}.}
\label{SI-tab:full_cold_davis_1}
\begin{tabular*}{\textwidth}{@{\extracolsep{\fill}}llcc@{}}
\toprule
Method & Source & MSE $\downarrow$ & CI $\uparrow$ \\
\midrule
GraphDTA & T24 & $0.439$ & $0.770$ \\
MGPLI & T24 & $0.457$ & $0.792$ \\
GRA-DTA & T24 & $0.376$ & $0.827$ \\
GDilatedDTA & Z24 & $0.382$ & $0.810$ \\
GLFA & N22 & $0.453$ & $0.780$ \\
GEFA & N22 & $0.434$ & $0.795$ \\
FusionDTA & H25 & $0.364$ & $0.826$ \\
MGraphDTA & H25 & $0.359$ & $0.813$ \\
DMFF & H25 & $0.330$ & $\mathbf{0.840}$ \\
\midrule
Ricci-GraphDTA & This study & $0.413 \pm 0.034$ & $0.796 \pm 0.033$ \\
\textbf{CurvFlow-DTA} & This study & $\mathbf{0.300 \pm 0.029}$ & $0.833 \pm 0.010$ \\
\bottomrule
\end{tabular*}
\end{table*}

\begin{table*}[p]
\centering
\small
\caption{Complete Davis dual-cold comparison. Sources are defined in Table~\ref{SI-tab:baseline_sources}.}
\label{SI-tab:full_cold_davis_2}
\begin{tabular*}{\textwidth}{@{\extracolsep{\fill}}llcc@{}}
\toprule
Method & Source & MSE $\downarrow$ & CI $\uparrow$ \\
\midrule
GraphDTA & T24 & $0.656$ & $0.607$ \\
MGPLI & T24 & $0.654$ & $0.556$ \\
GRA-DTA & T24 & $0.560$ & $0.690$ \\
GDilatedDTA & Z24 & $0.718$ & $0.633$ \\
GLFA & N22 & $1.144$ & $0.636$ \\
GEFA & N22 & $0.990$ & $0.639$ \\
FusionDTA & H25 & $0.876$ & $0.645$ \\
MGraphDTA & H25 & $0.874$ & $0.636$ \\
DMFF & H25 & $0.759$ & $0.655$ \\
\midrule
Ricci-GraphDTA & This study & $0.616 \pm 0.091$ & $0.696 \pm 0.088$ \\
\textbf{CurvFlow-DTA} & This study & $\mathbf{0.492 \pm 0.033}$ & $\mathbf{0.723 \pm 0.047}$ \\
\bottomrule
\end{tabular*}
\end{table*}

\begin{table*}[p]
\centering
\small
\caption{Complete KIBA cold-drug comparison. Sources are defined in Table~\ref{SI-tab:baseline_sources}.}
\label{SI-tab:full_cold_kiba_0}
\begin{tabular*}{\textwidth}{@{\extracolsep{\fill}}llcc@{}}
\toprule
Method & Source & MSE $\downarrow$ & CI $\uparrow$ \\
\midrule
GraphDTA & T24 & $0.349$ & $0.761$ \\
MGPLI & T24 & $0.500$ & $0.725$ \\
GRA-DTA & T24 & $\mathbf{0.337}$ & $\mathbf{0.765}$ \\
FusionDTA & H25 & $0.429$ & $0.748$ \\
MGraphDTA & H25 & $0.425$ & $0.746$ \\
DMFF & H25 & $0.408$ & $0.753$ \\
\midrule
Ricci-GraphDTA & This study & $0.406 \pm 0.002$ & $0.726 \pm 0.001$ \\
\textbf{CurvFlow-DTA} & This study & $0.347 \pm 0.032$ & $0.748 \pm 0.013$ \\
\bottomrule
\end{tabular*}
\end{table*}

\begin{table*}[p]
\centering
\small
\caption{Complete KIBA cold-target comparison. Sources are defined in Table~\ref{SI-tab:baseline_sources}.}
\label{SI-tab:full_cold_kiba_1}
\begin{tabular*}{\textwidth}{@{\extracolsep{\fill}}llcc@{}}
\toprule
Method & Source & MSE $\downarrow$ & CI $\uparrow$ \\
\midrule
GraphDTA & T24 & $0.519$ & $0.651$ \\
MGPLI & T24 & $0.531$ & $0.673$ \\
GRA-DTA & T24 & $0.435$ & $0.692$ \\
FusionDTA & H25 & $0.439$ & $0.685$ \\
MGraphDTA & H25 & $0.435$ & $0.674$ \\
DMFF & H25 & $0.410$ & $\mathbf{0.748}$ \\
\midrule
Ricci-GraphDTA & This study & $0.462 \pm 0.101$ & $0.710 \pm 0.020$ \\
\textbf{CurvFlow-DTA} & This study & $\mathbf{0.383 \pm 0.015}$ & $0.746 \pm 0.006$ \\
\bottomrule
\end{tabular*}
\end{table*}

\begin{table*}[p]
\centering
\small
\caption{Complete KIBA dual-cold comparison. Sources are defined in Table~\ref{SI-tab:baseline_sources}.}
\label{SI-tab:full_cold_kiba_2}
\begin{tabular*}{\textwidth}{@{\extracolsep{\fill}}llcc@{}}
\toprule
Method & Source & MSE $\downarrow$ & CI $\uparrow$ \\
\midrule
GraphDTA & T24 & $0.473$ & $0.647$ \\
MGPLI & T24 & $0.640$ & $0.611$ \\
GRA-DTA & T24 & $\mathbf{0.441}$ & $0.657$ \\
FusionDTA & H25 & $0.587$ & $0.641$ \\
MGraphDTA & H25 & $0.590$ & $0.626$ \\
DMFF & H25 & $0.567$ & $0.667$ \\
\midrule
Ricci-GraphDTA & This study & $0.534 \pm 0.016$ & $0.590 \pm 0.009$ \\
\textbf{CurvFlow-DTA} & This study & $0.445 \pm 0.059$ & $\mathbf{0.674 \pm 0.022}$ \\
\bottomrule
\end{tabular*}
\end{table*}

\clearpage

\putbib[references]

\end{bibunit}

\begin{thebibliography}{39}
\providecommand{\natexlab}[1]{#1}
\providecommand{\url}[1]{\texttt{#1}}
\expandafter\ifx\csname urlstyle\endcsname\relax
  \providecommand{\doi}[1]{doi: #1}\else
  \providecommand{\doi}{doi: \begingroup \urlstyle{rm}\Url}\fi

\bibitem[Alon and Yahav(2021)]{alon2021bottleneck}
U.~Alon and E.~Yahav.
\newblock On the bottleneck of graph neural networks and its practical
  implications.
\newblock In \emph{International Conference on Learning Representations
  (ICLR)}, 2021.

\bibitem[Bai et~al.(2026)Bai, Liu, and Lai]{bai2026weightedricciflow}
S.~Bai, S.~Liu, and X.~Lai.
\newblock The weighted {Forman} and {Lin--Lu--Yau Ricci} flow on graphs.
\newblock \emph{arXiv preprint arXiv:2601.02673}, 2026.

\bibitem[Bal et~al.(2023)Bal, Xiao, and Wang]{bal2023pgraphdta}
R.~Bal, Y.~Xiao, and W.~Wang.
\newblock {PGraphDTA}: Improving drug target interaction prediction using
  protein language models and contact maps.
\newblock \emph{arXiv preprint arXiv:2310.04017}, 2023.

\bibitem[Bemis and Murcko(1996)]{bemis1996frameworks}
G.~W. Bemis and M.~A. Murcko.
\newblock The properties of known drugs. 1. molecular frameworks.
\newblock \emph{Journal of Medicinal Chemistry}, 39\penalty0 (15):\penalty0
  2887--2893, 1996.

\bibitem[Chen et~al.(2025)Chen, Deng, Wang, Chen, and
  Zheng]{chen2025graphneuralricciflow}
J.~Chen, B.~Deng, Z.~Wang, C.~Chen, and Z.~Zheng.
\newblock Graph neural {Ricci} flow: Evolving feature from a curvature
  perspective.
\newblock In \emph{International Conference on Learning Representations
  (ICLR)}, 2025.

\bibitem[Chen et~al.(2020)Chen, Tan, Wang, Zhong, Liu, Yang, Luo, Chen, Jiang,
  and Zheng]{chen2020transformercpi}
L.~Chen, X.~Tan, D.~Wang, F.~Zhong, X.~Liu, T.~Yang, X.~Luo, K.~Chen, H.~Jiang,
  and M.~Zheng.
\newblock {TransformerCPI}: improving compound--protein interaction prediction
  by sequence-based deep learning with self-attention mechanism and label
  reversal experiments.
\newblock \emph{Bioinformatics}, 36\penalty0 (16):\penalty0 4406--4414, 2020.

\bibitem[Chen et~al.(2024)Chen, Wan, Li, He, Wei, and
  Han]{chen2024graphcurvatureflow}
Y.~Chen, Z.~Wan, Y.~Li, X.~He, X.~Wei, and J.~Han.
\newblock Graph curvature flow-based masked attention.
\newblock \emph{Journal of Chemical Information and Modeling}, 64\penalty0
  (21):\penalty0 8153--8163, 2024.

\bibitem[Davis et~al.(2011)Davis, Hunt, Herrgard, Ciceri, Wodicka, Pallares,
  Hocker, Treiber, and Zarrinkar]{davis2011comprehensive}
M.~I. Davis, J.~P. Hunt, S.~Herrgard, P.~Ciceri, L.~M. Wodicka, G.~Pallares,
  M.~Hocker, D.~K. Treiber, and P.~P. Zarrinkar.
\newblock Comprehensive analysis of kinase inhibitor selectivity.
\newblock \emph{Nature Biotechnology}, 29\penalty0 (11):\penalty0 1046--1051,
  2011.

\bibitem[Forman(2003)]{forman2003bochner}
R.~Forman.
\newblock Bochner's method for cell complexes and combinatorial ricci
  curvature.
\newblock \emph{Discrete \& Computational Geometry}, 29\penalty0 (3):\penalty0
  323--374, 2003.

\bibitem[He et~al.(2023)He, Chen, and Chen]{he2023nhgnndta}
H.~He, G.~Chen, and C.~Y.-C. Chen.
\newblock {NHGNN-DTA}: a node-adaptive hybrid graph neural network for
  interpretable drug--target binding affinity prediction.
\newblock \emph{Bioinformatics}, 39\penalty0 (6):\penalty0 btad355, 2023.

\bibitem[He et~al.(2025)He, Chen, Tang, and Chen]{he2025dmff}
H.~He, G.~Chen, Z.~Tang, and C.~Y.-C. Chen.
\newblock Dual modality feature fused neural network integrating binding site
  information for drug target affinity prediction.
\newblock \emph{npj Digital Medicine}, 8\penalty0 (1):\penalty0 67, 2025.

\bibitem[He et~al.(2017)He, Heidemeyer, Ban, Cherkasov, and
  Ester]{he2017simboost}
T.~He, M.~Heidemeyer, F.~Ban, A.~Cherkasov, and M.~Ester.
\newblock {SimBoost}: a read-across approach for predicting drug--target
  binding affinities using gradient boosting machines.
\newblock \emph{Journal of Cheminformatics}, 9\penalty0 (1):\penalty0 24, 2017.

\bibitem[Hu et~al.(2020)Hu, Liu, Gomes, Zitnik, Liang, Pande, and
  Leskovec]{hu2020strategies}
W.~Hu, B.~Liu, J.~Gomes, M.~Zitnik, P.~Liang, V.~Pande, and J.~Leskovec.
\newblock Strategies for pre-training graph neural networks.
\newblock In \emph{International Conference on Learning Representations
  (ICLR)}, 2020.

\bibitem[Jiang et~al.(2020)Jiang, Li, Zhang, Wang, Wang, Yuan, and
  Wei]{jiang2020dgraphdta}
M.~Jiang, Z.~Li, S.~Zhang, S.~Wang, X.~Wang, Q.~Yuan, and Z.~Wei.
\newblock Drug--target affinity prediction using graph neural network and
  contact maps.
\newblock \emph{RSC Advances}, 10\penalty0 (35):\penalty0 20701--20712, 2020.

\bibitem[Lin et~al.(2023)Lin, Akin, Rao, Hie, Zhu, Lu, Smetanin, Verkuil,
  Kabeli, Shmueli, dos Santos~Costa, Fazel-Zarandi, Sercu, Candido, and
  Rives]{lin2023esm2}
Z.~Lin, H.~Akin, R.~Rao, B.~Hie, Z.~Zhu, W.~Lu, N.~Smetanin, R.~Verkuil,
  O.~Kabeli, Y.~Shmueli, A.~dos Santos~Costa, M.~Fazel-Zarandi, T.~Sercu,
  S.~Candido, and A.~Rives.
\newblock Evolutionary-scale prediction of atomic-level protein structure with
  a language model.
\newblock \emph{Science}, 379\penalty0 (6637):\penalty0 1123--1130, 2023.

\bibitem[Nguyen et~al.(2021)Nguyen, Le, Quinn, Nguyen, Le, and
  Venkatesh]{nguyen2021graphdta}
T.~Nguyen, H.~Le, T.~P. Quinn, T.~Nguyen, T.~D. Le, and S.~Venkatesh.
\newblock {GraphDTA}: predicting drug--target binding affinity with graph
  neural networks.
\newblock \emph{Bioinformatics}, 37\penalty0 (8):\penalty0 1140--1147, 2021.

\bibitem[Nguyen et~al.(2022)Nguyen, Nguyen, Le, and Tran]{nguyen2022gefa}
T.~M. Nguyen, T.~Nguyen, T.~M. Le, and T.~Tran.
\newblock {GEFA}: Early fusion approach in drug-target affinity prediction.
\newblock \emph{IEEE/ACM Transactions on Computational Biology and
  Bioinformatics}, 19\penalty0 (2):\penalty0 718--728, 2022.

\bibitem[Ni et~al.(2019)Ni, Lin, Luo, and Gao]{ni2019community}
C.-C. Ni, Y.-Y. Lin, F.~Luo, and J.~Gao.
\newblock Community detection on networks with {Ricci} flow.
\newblock \emph{Scientific Reports}, 9\penalty0 (1):\penalty0 9984, 2019.

\bibitem[Ouyang et~al.(2025)Ouyang, Feng, Cui, Li, Zhang, and
  Wang]{ouyang2025pmmr}
X.~Ouyang, Y.~Feng, C.~Cui, Y.~Li, L.~Zhang, and H.~Wang.
\newblock Improving generalizability of drug--target binding prediction by
  pre-trained multi-view molecular representations.
\newblock \emph{Bioinformatics}, 41\penalty0 (1):\penalty0 btaf002, 2025.

\bibitem[{\"O}zt{\"u}rk et~al.(2018){\"O}zt{\"u}rk, {\"O}zg{\"u}r, and
  Ozkirimli]{ozturk2018deepdta}
H.~{\"O}zt{\"u}rk, A.~{\"O}zg{\"u}r, and E.~Ozkirimli.
\newblock {DeepDTA}: deep drug--target binding affinity prediction.
\newblock \emph{Bioinformatics}, 34\penalty0 (17):\penalty0 i821--i829, 2018.

\bibitem[Pahikkala et~al.(2015)Pahikkala, Airola, Pietil{\"a}, Shakyawar,
  Szwajda, Tang, and Aittokallio]{pahikkala2015toward}
T.~Pahikkala, A.~Airola, S.~Pietil{\"a}, S.~Shakyawar, A.~Szwajda, J.~Tang, and
  T.~Aittokallio.
\newblock Toward more realistic drug--target interaction predictions.
\newblock \emph{Briefings in Bioinformatics}, 16\penalty0 (2):\penalty0
  325--337, 2015.

\bibitem[Rakib et~al.(2026)Rakib, Alamin, Li, Mamun, Gobena, and
  Ren]{rakib2026kanpmdta}
M.~D. Y.~K. Rakib, M.~H. Alamin, J.~Li, S.~S. Mamun, K.~A. Gobena, and S.~Ren.
\newblock {KANPM-DTA}: improving drug--target affinity prediction with
  {Kolmogorov}--{Arnold} networks and pretrained models.
\newblock \emph{Briefings in Bioinformatics}, 27\penalty0 (2):\penalty0
  bbag112, 2026.

\bibitem[Rives et~al.(2021)Rives, Meier, Sercu, Goyal, Lin, Liu, Guo, Ott,
  Zitnick, Ma, and Fergus]{rives2021biological}
A.~Rives, J.~Meier, T.~Sercu, S.~Goyal, Z.~Lin, J.~Liu, D.~Guo, M.~Ott, C.~L.
  Zitnick, J.~Ma, and R.~Fergus.
\newblock Biological structure and function emerge from scaling unsupervised
  learning to 250 million protein sequences.
\newblock \emph{Proceedings of the National Academy of Sciences}, 118\penalty0
  (15):\penalty0 e2016239118, 2021.

\bibitem[Sreejith et~al.(2016)Sreejith, Mohanraj, Jost, Saucan, and
  Samal]{sreejith2016forman}
R.~P. Sreejith, K.~Mohanraj, J.~Jost, E.~Saucan, and A.~Samal.
\newblock {Forman} curvature for complex networks.
\newblock \emph{Journal of Statistical Mechanics: Theory and Experiment},
  2016\penalty0 (6):\penalty0 063206, 2016.

\bibitem[Steinegger and S{\"o}ding(2017)]{steinegger2017mmseqs2}
M.~Steinegger and J.~S{\"o}ding.
\newblock {MMseqs2} enables sensitive protein sequence searching for the
  analysis of massive data sets.
\newblock \emph{Nature Biotechnology}, 35\penalty0 (11):\penalty0 1026--1028,
  2017.

\bibitem[Tang et~al.(2014)Tang, Szwajda, Shakyawar, Xu, Hintsanen, Wennerberg,
  and Aittokallio]{tang2014making}
J.~Tang, A.~Szwajda, S.~Shakyawar, T.~Xu, P.~Hintsanen, K.~Wennerberg, and
  T.~Aittokallio.
\newblock Making sense of large-scale kinase inhibitor bioactivity data sets: a
  comparative and integrative analysis.
\newblock \emph{Journal of Chemical Information and Modeling}, 54\penalty0
  (3):\penalty0 735--743, 2014.

\bibitem[Tang et~al.(2025)Tang, Zhao, and Wang]{tang2025llmdta}
W.~Tang, Q.~Zhao, and J.~Wang.
\newblock {LLMDTA}: improving cold-start prediction in drug--target affinity
  with biological {LLM}.
\newblock \emph{IEEE Transactions on Computational Biology and Bioinformatics},
  22\penalty0 (6):\penalty0 2398--2409, 2025.

\bibitem[Tang et~al.(2024)Tang, Lei, and Zhang]{tang2024gradta}
X.~Tang, X.~Lei, and Y.~Zhang.
\newblock Prediction of drug-target affinity using attention neural network.
\newblock \emph{International Journal of Molecular Sciences}, 25\penalty0
  (10):\penalty0 5126, 2024.

\bibitem[Topping et~al.(2022)Topping, Di~Giovanni, Chamberlain, Dong, and
  Bronstein]{topping2022oversquashing}
J.~Topping, F.~Di~Giovanni, B.~P. Chamberlain, X.~Dong, and M.~M. Bronstein.
\newblock Understanding over-squashing and bottlenecks on graphs via curvature.
\newblock In \emph{International Conference on Learning Representations
  (ICLR)}, 2022.

\bibitem[Wang et~al.(2022)Wang, Hu, Sun, Xu, Yu, Liu, and Cheng]{wang2022mgpli}
J.~Wang, J.~Hu, H.~Sun, M.~Xu, Y.~Yu, Y.~Liu, and L.~Cheng.
\newblock {MGPLI}: exploring multigranular representations for protein--ligand
  interaction prediction.
\newblock \emph{Bioinformatics}, 38\penalty0 (21):\penalty0 4859--4867, 2022.

\bibitem[Wang et~al.(2024)Wang, Xiao, Shang, and Peng]{wang2024cscodta}
J.~Wang, Y.~Xiao, X.~Shang, and J.~Peng.
\newblock Predicting drug--target binding affinity with cross-scale graph
  contrastive learning.
\newblock \emph{Briefings in Bioinformatics}, 25\penalty0 (1):\penalty0
  bbad516, 2024.

\bibitem[Weber et~al.(2017)Weber, Saucan, and Jost]{weber2017formanflow}
M.~Weber, E.~Saucan, and J.~Jost.
\newblock Characterizing complex networks with {Forman}--{Ricci} curvature and
  associated geometric flows.
\newblock \emph{Journal of Complex Networks}, 5\penalty0 (4):\penalty0
  527--550, 2017.

\bibitem[Wee and Xia(2021)]{wee2021fprc}
J.~Wee and K.~Xia.
\newblock Forman persistent {Ricci} curvature ({FPRC})-based machine learning
  models for protein--ligand binding affinity prediction.
\newblock \emph{Briefings in Bioinformatics}, 22\penalty0 (6):\penalty0
  bbab136, 2021.

\bibitem[Wu et~al.(2023)Wu, Chen, Cheng, and Xiong]{wu2023curvagn}
J.~Wu, H.~Chen, M.~Cheng, and H.~Xiong.
\newblock {CurvAGN}: Curvature-based adaptive graph neural networks for
  predicting protein-ligand binding affinity.
\newblock \emph{BMC Bioinformatics}, 24\penalty0 (1):\penalty0 378, 2023.

\bibitem[Xu et~al.(2019)Xu, Hu, Leskovec, and Jegelka]{xu2019gin}
K.~Xu, W.~Hu, J.~Leskovec, and S.~Jegelka.
\newblock How powerful are graph neural networks?
\newblock In \emph{International Conference on Learning Representations
  (ICLR)}, 2019.

\bibitem[Yang et~al.(2022)Yang, Zhong, Zhao, and Chen]{yang2022mgraphdta}
Z.~Yang, W.~Zhong, L.~Zhao, and C.~Y.-C. Chen.
\newblock {MGraphDTA}: deep multiscale graph neural network for explainable
  drug--target binding affinity prediction.
\newblock \emph{Chemical Science}, 13\penalty0 (3):\penalty0 816--833, 2022.

\bibitem[Yuan et~al.(2022)Yuan, Chen, and Chen]{yuan2022fusiondta}
W.~Yuan, G.~Chen, and C.~Y.-C. Chen.
\newblock {FusionDTA}: attention-based feature polymerizer and knowledge
  distillation for drug-target binding affinity prediction.
\newblock \emph{Briefings in Bioinformatics}, 23\penalty0 (1):\penalty0
  bbab506, 2022.

\bibitem[Zheng et~al.(2026)Zheng, Zhang, Zhang, Jin, and
  Zhang]{zheng2026riccigraphdta}
X.~Zheng, Z.~Zhang, X.~Zhang, N.~Jin, and Y.~Zhang.
\newblock {Ricci-GraphDTA}: A graph neural network integrating discrete {Ricci}
  curvature for drug--target affinity prediction.
\newblock \emph{Journal of Molecular Graphics and Modelling}, 142:\penalty0
  109170, 2026.

\bibitem[Zhou et~al.(2026)Zhou, Chen, Zhou, and Xiao]{zhou2026pcimdta}
Z.~Zhou, M.~Chen, W.~Zhou, and Q.~Xiao.
\newblock {PCIM-DTA}: pairwise conditional interaction modeling for
  drug--target affinity prediction under cold-start scenarios.
\newblock \emph{Bioinformatics}, 42\penalty0 (9):\penalty0 btag646, 2026.

\end{thebibliography}


\begin{thebibliography}{22}
\providecommand{\natexlab}[1]{#1}
\providecommand{\url}[1]{\texttt{#1}}
\expandafter\ifx\csname urlstyle\endcsname\relax
  \providecommand{\doi}[1]{doi: #1}\else
  \providecommand{\doi}{doi: \begingroup \urlstyle{rm}\Url}\fi

\bibitem[Bellon et~al.(2012)Bellon, Whittington, and Long]{pdb3u6h}
S.~F. Bellon, D.~A. Whittington, and A.~L. Long.
\newblock Crystal structure of {c-Met} in complex with pyrazolone inhibitor 26,
  2012.
\newblock \doi{10.2210/pdb3U6H/pdb}.
\newblock PDB accession 3U6H.

\bibitem[Bemis and Murcko(1996)]{bemis1996frameworks}
G.~W. Bemis and M.~A. Murcko.
\newblock The properties of known drugs. 1. molecular frameworks.
\newblock \emph{Journal of Medicinal Chemistry}, 39\penalty0 (15):\penalty0
  2887--2893, 1996.

\bibitem[Davis et~al.(2011)Davis, Hunt, Herrgard, Ciceri, Wodicka, Pallares,
  Hocker, Treiber, and Zarrinkar]{davis2011comprehensive}
M.~I. Davis, J.~P. Hunt, S.~Herrgard, P.~Ciceri, L.~M. Wodicka, G.~Pallares,
  M.~Hocker, D.~K. Treiber, and P.~P. Zarrinkar.
\newblock Comprehensive analysis of kinase inhibitor selectivity.
\newblock \emph{Nature Biotechnology}, 29\penalty0 (11):\penalty0 1046--1051,
  2011.

\bibitem[He et~al.(2025)He, Chen, Tang, and Chen]{he2025dmff}
H.~He, G.~Chen, Z.~Tang, and C.~Y.-C. Chen.
\newblock Dual modality feature fused neural network integrating binding site
  information for drug target affinity prediction.
\newblock \emph{npj Digital Medicine}, 8\penalty0 (1):\penalty0 67, 2025.

\bibitem[He et~al.(2017)He, Heidemeyer, Ban, Cherkasov, and
  Ester]{he2017simboost}
T.~He, M.~Heidemeyer, F.~Ban, A.~Cherkasov, and M.~Ester.
\newblock {SimBoost}: a read-across approach for predicting drug--target
  binding affinities using gradient boosting machines.
\newblock \emph{Journal of Cheminformatics}, 9\penalty0 (1):\penalty0 24, 2017.

\bibitem[Lin et~al.(2020)Lin, Zhao, Xiao, Quan, Wang, and Yu]{lin2020deepgs}
X.~Lin, K.~Zhao, T.~Xiao, Z.~Quan, Z.-J. Wang, and P.~S. Yu.
\newblock {DeepGS}: Deep representation learning of graphs and sequences for
  drug-target binding affinity prediction.
\newblock In \emph{Proceedings of the 24th European Conference on Artificial
  Intelligence (ECAI 2020)}, volume 325 of \emph{Frontiers in Artificial
  Intelligence and Applications}, pages 1301--1308. IOS Press, 2020.

\bibitem[Lin et~al.(2023)Lin, Akin, Rao, Hie, Zhu, Lu, Smetanin, Verkuil,
  Kabeli, Shmueli, dos Santos~Costa, Fazel-Zarandi, Sercu, Candido, and
  Rives]{lin2023esm2}
Z.~Lin, H.~Akin, R.~Rao, B.~Hie, Z.~Zhu, W.~Lu, N.~Smetanin, R.~Verkuil,
  O.~Kabeli, Y.~Shmueli, A.~dos Santos~Costa, M.~Fazel-Zarandi, T.~Sercu,
  S.~Candido, and A.~Rives.
\newblock Evolutionary-scale prediction of atomic-level protein structure with
  a language model.
\newblock \emph{Science}, 379\penalty0 (6637):\penalty0 1123--1130, 2023.

\bibitem[Mukherjee et~al.(2022)Mukherjee, Ghosh, and
  Basuchowdhuri]{mukherjee2022deepglstm}
S.~Mukherjee, M.~Ghosh, and P.~Basuchowdhuri.
\newblock {DeepGLSTM}: Deep graph convolutional network and {LSTM} based
  approach for predicting drug-target binding affinity.
\newblock In \emph{Proceedings of the 2022 SIAM International Conference on
  Data Mining (SDM)}, pages 729--737. SIAM, 2022.

\bibitem[Nguyen et~al.(2021)Nguyen, Le, Quinn, Nguyen, Le, and
  Venkatesh]{nguyen2021graphdta}
T.~Nguyen, H.~Le, T.~P. Quinn, T.~Nguyen, T.~D. Le, and S.~Venkatesh.
\newblock {GraphDTA}: predicting drug--target binding affinity with graph
  neural networks.
\newblock \emph{Bioinformatics}, 37\penalty0 (8):\penalty0 1140--1147, 2021.

\bibitem[Nguyen et~al.(2022)Nguyen, Nguyen, Le, and Tran]{nguyen2022gefa}
T.~M. Nguyen, T.~Nguyen, T.~M. Le, and T.~Tran.
\newblock {GEFA}: Early fusion approach in drug-target affinity prediction.
\newblock \emph{IEEE/ACM Transactions on Computational Biology and
  Bioinformatics}, 19\penalty0 (2):\penalty0 718--728, 2022.

\bibitem[{\"O}zt{\"u}rk et~al.(2018){\"O}zt{\"u}rk, {\"O}zg{\"u}r, and
  Ozkirimli]{ozturk2018deepdta}
H.~{\"O}zt{\"u}rk, A.~{\"O}zg{\"u}r, and E.~Ozkirimli.
\newblock {DeepDTA}: deep drug--target binding affinity prediction.
\newblock \emph{Bioinformatics}, 34\penalty0 (17):\penalty0 i821--i829, 2018.

\bibitem[Pahikkala et~al.(2015)Pahikkala, Airola, Pietil{\"a}, Shakyawar,
  Szwajda, Tang, and Aittokallio]{pahikkala2015toward}
T.~Pahikkala, A.~Airola, S.~Pietil{\"a}, S.~Shakyawar, A.~Szwajda, J.~Tang, and
  T.~Aittokallio.
\newblock Toward more realistic drug--target interaction predictions.
\newblock \emph{Briefings in Bioinformatics}, 16\penalty0 (2):\penalty0
  325--337, 2015.

\bibitem[RDKit()]{rdkitsoftware}
RDKit.
\newblock {RDKit}: Open-source cheminformatics.
\newblock \url{https://www.rdkit.org}, 2026.
\newblock Version 2026.03.2; accessed 15 September 2026.

\bibitem[Schulze-Gahmen et~al.(2005)Schulze-Gahmen, Meijer, and Kim]{pdb1w0x}
U.~Schulze-Gahmen, L.~Meijer, and S.-H. Kim.
\newblock Crystal structure of human {CDK2} in complex with the inhibitor
  olomoucine, 2005.
\newblock \doi{10.2210/pdb1W0X/pdb}.
\newblock PDB accession 1W0X.

\bibitem[Steinegger and S{\"o}ding(2017)]{steinegger2017mmseqs2}
M.~Steinegger and J.~S{\"o}ding.
\newblock {MMseqs2} enables sensitive protein sequence searching for the
  analysis of massive data sets.
\newblock \emph{Nature Biotechnology}, 35\penalty0 (11):\penalty0 1026--1028,
  2017.

\bibitem[Tang et~al.(2014)Tang, Szwajda, Shakyawar, Xu, Hintsanen, Wennerberg,
  and Aittokallio]{tang2014making}
J.~Tang, A.~Szwajda, S.~Shakyawar, T.~Xu, P.~Hintsanen, K.~Wennerberg, and
  T.~Aittokallio.
\newblock Making sense of large-scale kinase inhibitor bioactivity data sets: a
  comparative and integrative analysis.
\newblock \emph{Journal of Chemical Information and Modeling}, 54\penalty0
  (3):\penalty0 735--743, 2014.

\bibitem[Tang et~al.(2024)Tang, Lei, and Zhang]{tang2024gradta}
X.~Tang, X.~Lei, and Y.~Zhang.
\newblock Prediction of drug-target affinity using attention neural network.
\newblock \emph{International Journal of Molecular Sciences}, 25\penalty0
  (10):\penalty0 5126, 2024.

\bibitem[Wang et~al.(2022)Wang, Hu, Sun, Xu, Yu, Liu, and Cheng]{wang2022mgpli}
J.~Wang, J.~Hu, H.~Sun, M.~Xu, Y.~Yu, Y.~Liu, and L.~Cheng.
\newblock {MGPLI}: exploring multigranular representations for protein--ligand
  interaction prediction.
\newblock \emph{Bioinformatics}, 38\penalty0 (21):\penalty0 4859--4867, 2022.

\bibitem[Yang et~al.(2022)Yang, Zhong, Zhao, and Chen]{yang2022mgraphdta}
Z.~Yang, W.~Zhong, L.~Zhao, and C.~Y.-C. Chen.
\newblock {MGraphDTA}: deep multiscale graph neural network for explainable
  drug--target binding affinity prediction.
\newblock \emph{Chemical Science}, 13\penalty0 (3):\penalty0 816--833, 2022.

\bibitem[Yuan et~al.(2022)Yuan, Chen, and Chen]{yuan2022fusiondta}
W.~Yuan, G.~Chen, and C.~Y.-C. Chen.
\newblock {FusionDTA}: attention-based feature polymerizer and knowledge
  distillation for drug-target binding affinity prediction.
\newblock \emph{Briefings in Bioinformatics}, 23\penalty0 (1):\penalty0
  bbab506, 2022.

\bibitem[Zhang et~al.(2024)Zhang, Zeng, Chen, Chen, and
  Li]{zhang2024gdilateddta}
L.~Zhang, W.~Zeng, J.~Chen, J.~Chen, and K.~Li.
\newblock {GDilatedDTA}: Graph dilation convolution strategy for drug target
  binding affinity prediction.
\newblock \emph{Biomedical Signal Processing and Control}, 92:\penalty0 106110,
  2024.

\bibitem[Zheng et~al.(2026)Zheng, Zhang, Zhang, Jin, and
  Zhang]{zheng2026riccigraphdta}
X.~Zheng, Z.~Zhang, X.~Zhang, N.~Jin, and Y.~Zhang.
\newblock {Ricci-GraphDTA}: A graph neural network integrating discrete {Ricci}
  curvature for drug--target affinity prediction.
\newblock \emph{Journal of Molecular Graphics and Modelling}, 142:\penalty0
  109170, 2026.

\end{thebibliography}
\end{document}